# Enhancing Research Idea Generation through Combinatorial Innovation and Multi-Agent Iterative Search Strategies

Shuai Chen and Chengzhi Zhang [*]

Department of Information Management, Nanjing University of Science and Technology, Nanjing, 210094 China
shuaichen@njust.edu.cn, zhangcz@njust.edu.cn

**Abstract.** Scientific progress depends on the continual generation of innovative research ideas. However, the rapid growth of scientific literature has greatly increased the cost of knowledge filtering, making it harder for researchers to identify novel directions. Although existing large language model (LLM)-based methods show promise in research idea generation, the ideas they produce are often repetitive and lack depth. To address this issue, this study proposes a multi-agent iterative planning search strategy inspired by combinatorial innovation theory. The framework combines iterative knowledge search with an LLM-based multi-agent system to generate, evaluate, and refine research ideas through repeated interaction, with the goal of improving idea diversity and novelty. Experiments in the natural language processing domain show that the proposed method outperforms state-of-the-art baselines in both diversity and novelty. Further comparison with ideas derived from top-tier machine learning conference papers indicates that the quality of the generated ideas falls between that of accepted and rejected papers. These results suggest that the proposed framework is a promising approach for supporting high-quality research idea generation. The source code and dataset used in this paper are publicly available on Github repository: https://github.com/ChenShuai00/MAGenIdeas. The demo is available at https://huggingface.co/spaces/cshuai20/MAGenIdeas.

## 1 Introduction

Over the past few decades, the volume of scientific literature has grown exponentially, reflecting the rapid expansion of research activities and the continuous advancement of science and technology. While this growth accelerates knowledge accumulation, it also imposes substantial cognitive and temporal costs on researchers who must continuously screen and interpret large bodies of literature. As the number of publications increases, identifying genuinely novel research opportunities becomes increasingly difficult. This challenge not only contributes to redundant scientific activities (Larivière et al., 2008) but also leads to inefficient allocation of research resources, ultimately raising the difficulty of conducting innovative research.

---

[*] Corresponding author.

Recently, large language models (LLMs) have demonstrated remarkable capabilities across a wide range of complex tasks, including mathematical reasoning (Yan et al., 2023), information retrieval (Ajith et al., 2024), and solving research problems through code generation (Lu et al., 2024; Schmidgall et al., 2025; Yuan et al., 2025). More importantly, emerging studies suggest that LLMs can assist researchers in generating scientific hypotheses and research ideas. However, existing work has also shown that ideas generated by LLMs often contain substantial conceptual redundancy and repeated research directions (Si et al., 2024). As a result, researchers still need to invest considerable effort in filtering and refining the generated ideas, which limits the practical applicability of LLM-based research ideation systems.

To address this issue, several studies have explored methods for guiding LLMs in generating research ideas. Some approaches retrieve relevant literature and incorporate it into prompts to stimulate idea generation (Guo et al., 2025; Lu et al., 2024). Others employ retrieval-augmented generation (RAG) frameworks to dynamically integrate external knowledge during idea construction (Si et al., 2024). More recent work introduces structured knowledge representations, such as scientific knowledge graphs or entity co-occurrence networks, to improve idea diversity (Baek et al., 2025; Gu & Krenn, 2024). In addition, iterative knowledge search strategies have been proposed to reduce redundancy in generated ideas (Hu et al., 2025).

Despite these advances, existing approaches still face several fundamental limitations. First, many systems rely on single-agent or single-role prompting strategies, which restrict the diversity of perspectives involved in idea exploration and evaluation. Consequently, even when retrieval or iterative planning is introduced, the generation process may still be dominated by a single interpretive viewpoint, leading to perspective bias and path dependency in the generated ideas. Second, although methods such as NOVA (Hu et al., 2025) improve the efficiency of knowledge retrieval through iterative search, they do not explicitly address how retrieved knowledge elements can be recombined across heterogeneous viewpoints to stimulate novel conceptual combinations. Third, current frameworks rarely simulate the collaborative nature of scientific discovery, in which research ideas are continuously refined through interactions among researchers with different expertise and cognitive perspectives.

These limitations can be better understood through the lens of combinatorial innovation theory. Schumpeter (1964) proposed that innovation fundamentally arises from novel recombinations of existing elements, such as technologies, resources, or knowledge components. In the context of scientific discovery, research ideas are often generated by integrating concepts, methods, and datasets originating from different domains (Schumpeter, 1964). Building on this perspective, recent studies conceptualize knowledge creation as a dynamic process of searching, selecting, and reorganizing existing knowledge elements (Ke et al., 2026). From this viewpoint, generating novel research ideas requires not only retrieving relevant knowledge but also recombining knowledge elements through multiple heterogeneous perspectives.

Inspired by these insights, this paper proposes a multi-agent iterative planning and search framework for automated research idea generation. The proposed approach first generates initial research ideas based on a target paper and its references, and then iteratively expands the knowledge space through planned literature search. Meanwhile, a virtual team of agents is constructed using background information derived from the authors of the target paper. Each agent represents a distinct academic perspective and independently evaluates, critiques, and refines research ideas during the iterative process. Through the interaction of multiple agents and continuous knowledge integration, the framework enables more diverse and innovative recombinations of knowledge elements.

Experiments conducted on a dataset from the natural language processing (NLP) domain demonstrate that the proposed approach significantly improves the diversity, novelty, and quality of generated research ideas compared with baseline methods. Furthermore, by comparing generated ideas with accepted and rejected submissions from the International Conference on Learning Representations 2025 (ICLR 2025), we show that the quality of generated ideas falls between rejected and accepted papers, indicating that the proposed framework can produce research ideas with meaningful academic potential.

To guide the investigation, this study addresses the following research questions:

**RQ1:** Can the multi-agent iterative planning and search framework improve the diversity and novelty of research ideas generated by LLMs?

**RQ2:** Can combinatorial innovation theory provide a theoretical foundation for guiding LLM-based research idea generation?

This study makes three main contributions.

First, we propose a multi-agent iterative planning and search framework for automated research idea generation. The framework integrates cross-domain knowledge planning with collaborative multi-agent reasoning to expand the exploration space of LLMs.

Second, we introduce combinatorial innovation theory as the theoretical foundation for guiding knowledge recombination during the idea generation process. This provides a principled explanation of how structured knowledge recombination can enhance the diversity and novelty of generated research ideas.

Third, we conduct extensive experiments to evaluate the proposed framework. The results demonstrate that the proposed approach consistently outperforms existing baselines in terms of diversity, novelty, and quality of generated research ideas.

## 2 Related work

This section reviews related work from two aspects: research idea generation using LLMs, and prompt engineering to stimulate the logical reasoning capabilities of LLMs.

## 2.1 Generating Research Ideas Using LLMs

Recent studies have explored various approaches to generating research ideas using LLMs. One line of work focuses on grounding idea generation in external knowledge sources. For example, some studies retrieve relevant papers based on research topics and incorporate them into prompts to stimulate idea generation (Guo et al., 2025; Lu et al., 2024). Others employ retrieval-augmented generation frameworks to dynamically integrate external knowledge during the idea construction process (Si et al., 2024). More structured approaches further utilize scientific knowledge graphs or entity co-occurrence networks to guide idea generation through structured knowledge representations (Baek et al., 2025; Gu & Krenn, 2024).

Another line of research aims to improve the idea generation process through human involvement or model training. IdeaSynth (Pu et al., 2025) incorporates human feedback into the generation loop and shows that human-AI collaboration can outperform single-model baselines. Similarly, (Li et al., 2024) enhance the feasibility and novelty of generated ideas through supervised fine-tuning and reinforcement learning.
More recently, researchers have explored multi-agent collaboration to simulate scientific teamwork. For instance, Virtual-Scientists (VIRSCI) (Su et al., 2025) models scientific collaboration by allowing multiple LLM agents to interact and refine research ideas collectively.

Although recent work such as NOVA (Hu et al., 2025) introduces iterative planning and knowledge search to reduce redundancy in LLM-generated research ideas, several limitations remain. First, NOVA operates under a single-agent and single-role prompting paradigm, which restricts the diversity of perspectives involved in idea evaluation and refinement and makes the generation process susceptible to perspective bias and path dependency. Second, NOVA mainly focuses on improving the efficiency of knowledge retrieval and search planning but does not explicitly address how retrieved knowledge elements can be recombined across heterogeneous viewpoints. From the perspective of combinatorial innovation, retrieval alone cannot guarantee novel recombinations if knowledge interpretation is guided by a single evaluative perspective. Third, existing approaches generally lack mechanisms to simulate the collaborative nature of scientific discovery, in which ideas are iteratively refined through interactions among researchers with different expertise.

In contrast, our work introduces a multi-agent iterative planning and search framework that incorporates differentiated agent perspectives derived from real-world author information. By combining knowledge planning with multi-agent evaluation and refinement, the proposed method expands the combinatorial search space and mitigates single-perspective bias, thereby enabling more diverse and novel research idea generation.

## 2.2 Reasoning and Multi-Agent LLMs

Prompting strategies have been widely studied as an effective means of enhancing the reasoning capabilities of LLMs. Chain-of-Thought (CoT) prompting (Wei et al., 2022) enables LLMs to solve complex tasks by explicitly generating intermediate reasoning steps, improving both interpretability and accuracy. Building upon this idea, Zero-Shot Chain-of-Thought prompting (Kojima et al., 2022) demonstrates that simple reasoning cues such as "Let's think step by step" can elicit reasoning chains without requiring manually designed examples.

Subsequent research has explored more structured reasoning strategies. Auto-CoT (Zhang et al., 2022) automatically constructs reasoning demonstrations to reduce the cost of manual prompt design. Self-consistency (X. Wang et al., 2022) improves reasoning reliability by sampling multiple reasoning paths and selecting the most consistent answer. Least-to-Most prompting (Zhou et al., 2022) introduces a planning-based approach that decomposes complex problems into sequential sub-problems. Similarly, the Tree-of-Thought (ToT) framework (Yao et al., 2023) enables LLMs to explore multiple reasoning branches and perform self-evaluation during the reasoning process.

However, reasoning generated by a single LLM can still suffer from biases and limited perspectives, which may affect the reliability of generated results (Liusie et al., 2023; Peiyi Wang et al., 2023). To address these limitations, recent studies have introduced multi-agent frameworks that simulate collaborative reasoning processes among multiple LLM agents. These approaches employ structured interaction mechanisms such as role-playing (N. Wu et al., 2023), debate (Chan et al., 2023), and voting (Zhu et al., 2024), allowing different agents to provide complementary perspectives and improving the robustness of model reasoning.

Building upon these developments, this study integrates role-based multi-agent collaboration with iterative planning to address the complex task of automated research idea generation.

## 2.3 Theoretical Foundations of Scientific Innovation

The generation of novel research ideas has long been studied in the literature on scientific and technological innovation. A foundational perspective is Schumpeter's combinatorial view of innovation (Schumpeter, 1964), which conceptualizes novelty as emerging from the recombination of existing knowledge elements into new configurations. This perspective has influenced science-of-science studies that operationalize innovation through atypical knowledge recombination. For example, (Uzzi et al., 2013) show that high-impact research often involves atypical combinations of prior knowledge, while (Lee et al., 2015) further examine how novelty and impact interact in scientific teamwork.

Complementary perspectives highlight the mechanisms through which such recombination processes may generate innovative solutions. The Theory of Inventive Problem Solving (TRIZ) (Altshuller, 1984) emphasizes systematic reasoning strategies, such as identifying contradictions and restructuring problem representations, to derive inventive solutions. In addition, disruptive innovation theory (Christensen, 1997) suggests that transformative advances may emerge when research departs from dominant trajectories rather than extending them incrementally.

Building on these perspectives, prior studies have proposed several indicators for measuring scientific novelty and innovation. Combinational approaches measure atypical knowledge recombination (Lee et al., 2015; Uzzi et al., 2013), network-based approaches analyze structural changes in knowledge systems, such as disruptiveness and paradigm shifts (Funk & Owen-Smith, 2017; Prabhakaran et al., 2015; Prabhakaran et al., 2018; L. Wu et al., 2019), and semantic approaches estimate conceptual novelty using textual representations and embeddings (Shibayama et al., 2021; Yan & Fan, 2024). Together, these perspectives provide a multidimensional view of scientific innovation (Wang et al., 2017; Bu et al., 2021).

In this study, combinatorial innovation serves as the primary theoretical foundation. Our framework operationalizes this principle by enabling LLM agents to retrieve, reinterpret, and recombine knowledge from existing literature. The iterative planning mechanism reflects structured reasoning ideas related to TRIZ, while the multi-agent design encourages exploration beyond dominant research trajectories. However, since generated ideas lack citation histories, novelty and diversity are evaluated at the conceptual level using semantic representations.

## 3  Methods

This section details the complete workflow of the multi-agent iterative planning and search method. It primarily consists of four key steps: dataset construction; initial research idea generation; iterative refinement of research ideas; and research idea abstract generation. The research framework of this paper is illustrated in Figure 1.

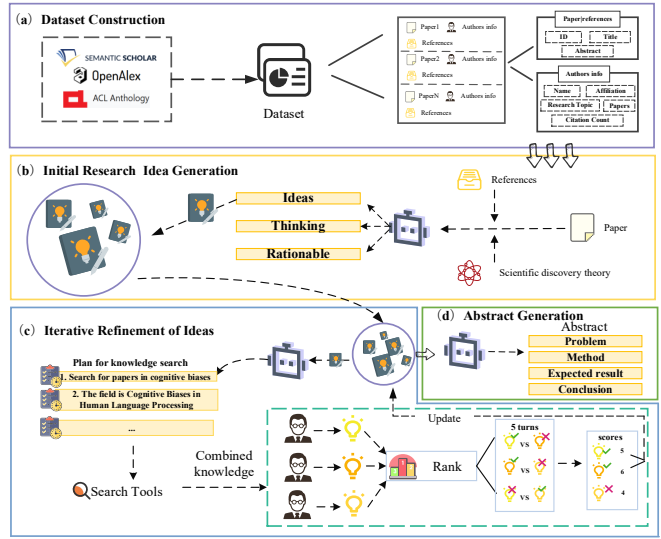

**Figure. 1.** Framework of this study

## 3.1 Dataset Construction

We select natural language processing (NLP) as the experimental domain for both methodological and practical reasons. First, NLP is a mature and highly active field with rapid publication cycles and substantial conceptual overlap across subfields, making it well suited for evaluating research idea generation, where redundancy and limited novelty are common challenges. Second, because large language models are primarily trained on textual data, NLP provides a natural setting for assessing their generative, reasoning, and evaluative capabilities while reducing confounding effects caused by modality mismatch. Third, the NLP community follows relatively standardized conventions in paper structure, methodological reporting, and evaluation, which facilitates controlled comparison between generated ideas and real conference submissions.

The dataset in this study serves two purposes: generating initial research ideas and constructing background knowledge for the multi-agent system. Accordingly, it must include target papers, their references, and metadata about the authors. Prior work has shown that high-quality source papers can substantially improve the quality of ideas generated by LLMs (Peng Wang et al., 2024). Guided by this finding, we select long papers from ACL 2024 as the initial corpus. ACL 2024 provides a recent and carefully curated snapshot of the NLP research frontier, while ACL, as a flagship venue, maintains high standards of originality, rigor, and reproducibility. We focus on long papers because they typically present more mature and complete research contributions, offering a denser and more coherent knowledge base for downstream idea generation. Restricting the dataset to a single flagship conference also reduces cross-venue variation in review standards, topical scope, and writing conventions, thereby improving the internal validity and interpretability of the experiments.

However, a single database is insufficient to meet the data requirements of this study. Therefore, we integrate the ACL Anthology[1], OpenAlex[2], and Semantic Scholar[3] as complementary data sources. In total, we initially collected 675 target papers and 22,647 references. After quality screening, the final dataset contains 144 target papers, 6,153 references, 953 author profiles, and 25,906 additional papers published by the corresponding authors.

The target paper dataset is constructed by combining multiple publicly available academic databases. ACL Anthology serves as the primary source for identifying candidate papers, while OpenAlex and Semantic Scholar are used only to enrich metadata, including references, citation counts, and author profiles. Specifically, all 675 target papers are drawn exclusively from the ACL 2024 long paper track; OpenAlex and Semantic Scholar do not introduce additional target papers. Because all target papers come from ACL 2024, a flagship NLP venue, the domain constraint is naturally satisfied. These external databases are queried using paper titles and DOIs rather than through additional topic-based filtering.

To integrate data across sources, we use paper titles and DOIs as unique identifiers when available. For each ACL paper, OpenAlex provides structured metadata such as references, citation counts, and institutional information, while Semantic Scholar supplements citation statistics and author publication records. When duplicate or conflicting fields arise, priority is assigned to ACL Anthology for paper identity, OpenAlex for reference relations, and Semantic Scholar for citation-related attributes.

---

[1] https://aclanthology.org/

[2] https://openalex.org/

[3] https://www.semanticscholar.org/

To ensure metadata completeness and sufficient bibliographic context for downstream processing, we apply deterministic filtering criteria to the initial set of 675 ACL 2024 long papers. A paper is retained only if it has at least 10 citations, includes no fewer than 20 references, and provides complete author information in at least one auxiliary database. These criteria are not used to judge academic quality, but to ensure that each paper contains sufficient citation, reference, and author metadata for knowledge retrieval, author-informed agent construction, and novelty evaluation. Papers with sparse bibliographic information often lead to incomplete reference graphs and unstable retrieval results, reducing the reliability and reproducibility of the pipeline. Similar citation-based filtering strategies have also been adopted in recent work on LLM-driven research idea generation (Hu et al., 2025). After filtering, 144 target papers are retained for subsequent experiments.

The final dataset includes paper titles and abstracts for target papers and their references, as well as author-related metadata such as research interests, institutional affiliations, publication counts, citation counts, and publication lists. To protect privacy, sensitive information such as author names is anonymized. All data collection procedures are reproducible through the public APIs of ACL Anthology, OpenAlex, and Semantic Scholar.

### 3.2 Initial Research Idea Generation

This paper first randomly selects a target paper to define the direction for research idea generation, which also determines the size of the virtual academic agent team. (Sadler-Smith, 2015)posits that the creative process can be divided into four stages: preparation, incubation, illumination, and verification. The purpose of initial research idea generation is to prepare and incubate for the subsequent iteration of genuine ideas by the virtual academic agent team. Therefore, drawing on the approach of Nova (Hu et al., 2025), this paper designs an initial idea generation module. Specifically, given a target paper (research idea direction), along with its references (background information for research ideas) and scientific discovery theories (research idea conception methods), these are input into a LLM to generate research ideas.

To enhance the scientific rigor and diversity of the initial research ideas, this paper selects ten scientific discovery methods to constrain and stimulate the generation process, guiding the LLM to produce innovative ideas from the input paper and its references. For example, drawing on Peirce's hypothetico-deductive method, it starts from facts and propositions, proposes a hypothesis or proposition as a premise, then performs logical reasoning based on the existing premises to derive conclusions; by combinatorially analyzing the relationships between premises, the truth value and validity of the conclusions can be determined.

Simultaneously, to simulate the process of research idea creation, this study utilizes the LLM to first comprehend the input target paper and its references, extract their core viewpoints, then select appropriate scientific discovery methods to generate research ideas, and provide the reasoning process involved, ensuring the interpretability of the results (Wei et al., 2022). Finally, 15 initial research ideas are generated to form a research idea pool, preparing for subsequent iterations.

To formalize the prompting process, this paper assumes $P$ represents the target paper, $L$ its references, $T$ the scientific method theory, and $R$ the generated research idea. The initial research idea generation can then be expressed as:

$$R = f(P, L, T) \quad (1)$$

where, $f$ represents the LLM, which utilizes its language understanding capabilities to generate research ideas. The prompts and examples for initial research idea generation can be found in Appendix Figure. 3 and Figure. A2.

### 3.3 Research Idea Iteration

Previous methods have predominantly relied on keyword-based searches or co-occurrence of entity concepts to incorporate external knowledge. However, these approaches exhibit significant limitations, such as inaccurate or overly broad search results, which hinder the ability of LLMs to engage in deep reasoning (Hu et al., 2025).

To effectively address these shortcomings, this study integrates planning principles into the knowledge search phase of research idea generation. Specifically, the LLM is utilized to meticulously plan and design knowledge search tasks, which are then executed sequentially using external academic search APIs. Ultimately, knowledge from diverse domains that is closely related to the research idea is combinatorially integrated into the LLM's prompts, providing more targeted and novel composite knowledge for idea generation. The prompt templates and examples for knowledge planning and search are provided in Appendix Figure. A3 and Figure. A4.

The virtual academic agent team constructed in this paper consists of multiple agents possessing the background knowledge of real scientists, denoted by $S = [s_1, s_2, ..., s_n]$, where $S$ represents the entire virtual academic agent team and $n$ is the team size. The background knowledge of the agents is derived from the author team information of the target paper. During the iterative generation process of research ideas, after acquiring new knowledge, each agent proposes its own research ideas and performs self-checking and scoring according to the best practice guidelines of AI conference reviews (e.g., ICLR and ACL) (Si et al., 2024). The scoring rubric is provided to each agent as part of the prompt context. The detailed scoring rubric can be found in Appendix Figure. A10.

The research ideas generated by each agent are evaluated for their creative quality through a Swiss System Tournament and a zero-shot LLM ranker. The ranker employs a pairwise comparison approach to determine which idea is superior. Each idea undergoes 5 rounds of comparison, receiving 1 point for each win. Practice has proven that this quality evaluation method is more advantageous compared to directly assigning scores (Si et al., 2024). Ultimately, ideas with a score of 5 points or above are selected as the final output of this iteration. Meanwhile, the negative comments recorded during the comparison process, along with the selected final ideas, proceed to the next iteration. The prompting process for each agent can be represented as:

$$R_i = f(R_t, K, B) \tag{2}$$

where $R_i$ represents the research idea generated by the *i-th* agent, $R_t$ denotes the research idea produced in the *t-th* round, $K$ is the new knowledge acquired through planned search, and $B$ represents the feedback on the research ideas generated in the *t-th* round. The prompts and examples for research idea generation can be found in Appendix Figure. A5 and Figure. A6, while the prompts for research idea comparison are provided in Appendix Figure. A7.

In each iteration, newly generated research ideas replace the old ones. Through this approach, the agents in this study are able to conduct more in-depth research exploration, significantly expanding the boundaries of the research search space.

### 3.4 Research Idea Abstract Generation

After *T* iterations, the final research ideas are established. In this process, the study draws on the summary generation method proposed by VIRSCI (Su et al., 2025). Specifically, the finalized research ideas are input into the LLM with a rigorously defined summary format (including aspects such as objectives and problems, methods, expected results, and conclusions), ensuring that the research ideas are presented in a detailed and structured manner. Additionally, since the summaries will subsequently be compared with reference paper abstracts for evaluation, outputting the research ideas in summary form is both practical and aligned with the assessment requirements. The prompt templates and examples for research idea summary generation can be found in Appendix Figure. A8 and Figure. A9.

## 4 Experiment and Results

### 4.1 Experimental Setup

**Large Language Model Configuration**: We implement the proposed framework in AgentScope (Gao et al., 2024). DeepSeek-V3 is used as the primary backbone model in the main experiments. In addition, to examine the cross-model applicability of the framework, we further instantiate the same framework with GPT-4o (2024-11-20) and qwen3-8b. Unless otherwise specified, all backbone models are run with their default decoding configurations in the corresponding deployment environments, and no additional hyperparameter tuning is applied to different agent roles. Specifically, DeepSeek-V3 is accessed through the DeepSeek API[4], GPT-4o through the OpenAI API[5], and qwen3-8b through the OpenRouter API [6] . In our implementation, neither the temperature nor the top-p parameter is manually specified. Therefore, all three models follow the default API sampling settings provided by their corresponding platforms, with temperature = 1.0 and top-p left at the provider default value. This design reduces confounding effects introduced by manual decoding adjustment and ensures that the comparison focuses primarily on framework-level differences rather than model-specific parameter tuning.

---

[4] https://platform.deepseek.com/
[5] https://openai.com/
[6] https://openrouter.ai/

**Baseline**: To evaluate the effectiveness of the proposed method, we compare it with state-of-the-art research idea generation approaches. The first baseline is AI-Researcher (Si et al., 2024), an end-to-end framework that retrieves relevant literature and prompts large language models to generate research ideas, while using pairwise comparison to rank idea quality. We also include NOVA (Hu et al., 2025), which introduces an iterative planning framework that guides knowledge retrieval and idea generation through planning-driven search, enabling broader exploration of the knowledge space and improving the novelty and diversity of generated ideas.

**Implementations Details**: We randomly selected 5 papers for each team size ranging from 2 to 8 members, resulting in a total of 35 papers. 525 initial research ideas were generated. Each baseline method also generated 5 sets of data, totaling 75 research ideas for evaluation. Evaluation was conducted based on metrics including the average proportion of high-scoring ideas, average novelty, and average diversity. During the experimental process, utilizing a multi-agent iterative planning search approach, after three rounds of iteration, a cumulative total of 2,027 research ideas were generated, with the first round producing 568 research ideas, the second round generating 656 research ideas, and the third round yielding 803 research ideas. Specifically, in each round, 8-person teams contributed an average of 126 research ideas, 7-person teams contributed 113, 6-person teams contributed 107, 5-person teams contributed 97, 4-person teams contributed 79, 3-person teams contributed 77, and 2-person teams contributed 75. The trend of the average number of ideas corresponding to each team size is shown in Figure. 2, clearly indicating that larger team sizes result in a greater number of research ideas filtered through the LLM self-scoring in each round.

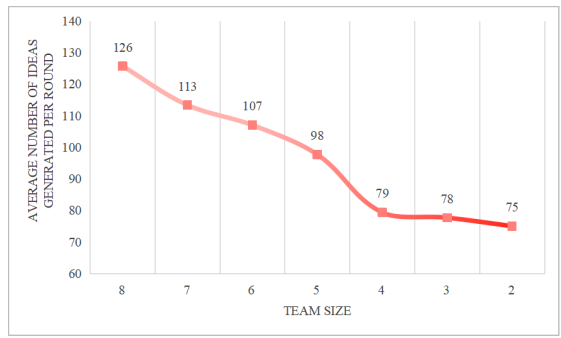

**Figure. 2.** Average Number of Ideas Generated per Team per Iteration

## 4.2 Automatic Evaluation

This study adopts the evaluation protocols of AI-Researcher (Si et al., 2024) and NOVA (Hu et al., 2025) to assess LLM-generated research ideas from three dimensions: quality score, diversity, and novelty. In addition, by comparing generated ideas with those derived from both accepted and rejected papers at ICLR 2025, we further evaluate their practical performance in a realistic academic setting.

The evaluation framework is grounded in the assumption that scientific novelty arises from conceptual differentiation from existing knowledge, whereas diversity reflects the breadth of conceptual exploration within a set of generated ideas. In bibliometric research, these properties are often measured using citation-based indicators that depend on long-term citation trajectories. However, because the generated ideas in this study have not entered the publication system and therefore have no citation histories, such indicators are not applicable. We therefore assess novelty and diversity at the conceptual level using semantic similarity and semantic distance: novelty is measured by the semantic similarity between generated ideas and existing literature, while diversity is measured by the semantic distance among generated ideas.

**Quality Score**: Utilizing the Swiss-system tournament and zero-shot LLM prompting method to evaluate the quality of research ideas, specifically, the ranking is determined through pairwise comparisons to judge which idea is superior. Each idea undergoes 5 rounds of comparison, with 1 point awarded for each winning comparison. This quality assessment method has been proven superior to direct comparison or scoring (Si et al., 2024). Ideas with scores greater than or equal to 5 are considered high-scoring ideas. The quality is ultimately measured by the proportion of high-scoring ideas, calculated using the following formula:

$$HighScoreRatio = \frac{\sum_{i}^{n} I(s_i > 4)}{n} \quad (3)$$

where $n$ is the number of generated research ideas, $s_i$ denotes the score of the $i$-th idea, and $I(s_i > 4)$ represents the indicator function that equals 1 when $s_i > 4$ and 0 otherwise.

**Novelty**: This paper employs a LLM to evaluate the novelty of generated ideas by examining the 10 most relevant papers (the method for retrieving the most relevant papers is consistent with the knowledge search approach described in Section 3.3). An idea is deemed novel if no paper is identified as containing similar concepts, using the all-MiniLM-L6-v2[7] model for embedding. A cosine similarity threshold greater than 0.5 indicates similarity. The novelty calculation formula is as follows:

$$Novelty = \frac{\sum_{i=1}^{n} I(max_j sim(a_i, r_{ij}) < \theta)}{n} \quad (4)$$

where $n$ is the number of generated research ideas, $sim(a_i, r_{ij})$ denotes the cosine similarity between idea $a_i$ and its relevant literature abstract $r_{ij}$, and $I(\cdot)$ represents the indicator function that returns 1 when the condition is true and 0 otherwise.

---

[7] https://huggingface.co/sentence-transformers/all-MiniLM-L6-v2

**Diversity**: Similar to (Hu et al., 2024; Si et al., 2024), the proportion of unique ideas is used to measure the diversity of generated research ideas. Specifically, the same similarity measurement method as used in the novelty metric is employed, with the duplication threshold set to 0.8. The diversity calculation formula is as follows:

$$Diversity = \frac{1}{n}\sum_{i=1}^{n} I\left(max_{j \neq i} sim(i,j) < threshold\right) \qquad (5)$$

where $n$ is the number of generated research ideas, $sim(i,j)$ represents the cosine similarity between idea $i$ and idea $j$, threshold denotes the similarity threshold, and $I(\cdot)$ is the indicator function that returns 1 when the condition is true and 0 otherwise.

**Comparative validation with top conference paper ideas**: We collected peer-review data for ICLR 2025 from OpenReview, including 3,704 accepted papers and 4,931 rejected papers. Among the accepted papers, 213 were oral presentations, 380 were spotlight presentations, and 3,111 were poster presentations.

To focus on the NLP domain, we used LLM-based text classification to identify NLP-related papers. This process yielded 842 accepted and 803 rejected NLP papers. To ensure a fair comparison, we excluded 72 controversial rejected papers whose average review scores exceeded the acceptance threshold, resulting in 731 rejected NLP papers. The numbers of NLP papers by presentation type and their average review scores are reported in Table 1.

OpenReview provides publicly accessible metadata for ICLR 2025, including final decisions, presentation types, and review scores. All data used in this study are publicly released by the conference organizers and can be accessed programmatically without authentication. We strictly complied with OpenReview's terms of use and did not access any private, anonymized, or restricted review content.

**Table 1.** The number of NLP papers by presentation type and the average review score

| Presentation type | Oral | Spotlight | Poster | Reject |
|---|---|---|---|---|
| Count | 61.0 | 88.0 | 693.0 | 731.0 |
| Mean | 7.759 | 7.325 | 6.230 | 4.667 |
| Std | 0.665 | 0.264 | 0.498 | 0.8850 |
| Min | 5.667 | 6.0 | 4.0 | 1.0 |
| Max | 9.0 | 8.0 | 7.0 | 5.833 |

To support fair comparison with generated content, we normalized all paper abstracts using the unified rewriting method described in Section 3.4. For evaluation, we combined cross-group Swiss-system pairing with zero-shot LLM-based pairwise comparison. In each round, generated ideas were paired with ideas from real papers based on current scores, with random pairing in the first round and score-based matching thereafter. Each pair was compared by an LLM along four dimensions—novelty, scientific rigor, clarity, and impact—and the better idea received 1 point. Scores were updated after each round, and the process was repeated for five rounds while avoiding rematches.

We considered three comparison settings: accepted versus rejected paper ideas, generated ideas versus accepted paper ideas, and generated ideas versus rejected paper ideas. By comparing average scores across the five rounds, we assessed the relative quality of generated research ideas in a realistic academic context.

## 4.3 Comparison with Baseline Methods

This section answers RQ1 through a controlled experimental comparison. To ensure that the observed performance differences arise from the generation strategies rather than from variations in evaluation data or initialization conditions, we align the experimental setup of all methods with the framework proposed in this study. Specifically, our method, AI-Researcher (Si et al., 2024), and NOVA (Hu et al., 2025) all sample papers from the same candidate paper pool, use the same fixed random seed for paper selection, and store the sampled subset to ensure full reproducibility. More importantly, all methods start from the same initial research ideas provided in the dataset, thereby ensuring identical input conditions across methods.

Under this unified setup, AI-Researcher follows its original two-stage pipeline of literature retrieval and literature-grounded idea generation, while NOVA guides knowledge retrieval and idea generation through a planning-driven search mechanism. Our method and the two baseline approaches are evaluated using the same sampled papers and initial ideas, ensuring a fair comparison.

The comparative results are shown in Figure 3 based on three metrics: diversity, novelty, and high-score ratio. Diversity reflects the breadth of explored research directions, novelty measures the extent to which generated ideas differ from existing literature, and high-score ratio indicates the proportion of ideas identified as high quality in the Swiss-system pairwise evaluation.

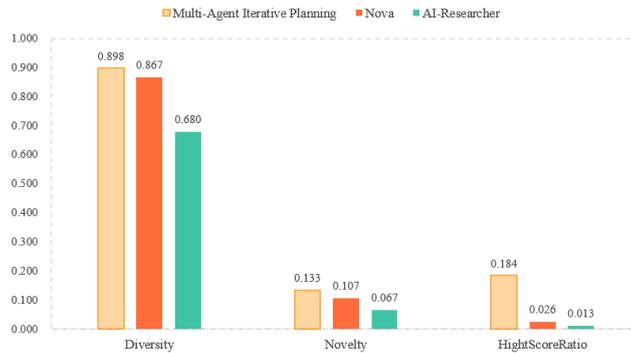

**Figure. 3.** Comparison with Baseline Methods

As shown in Figure 3, the proposed multi-agent iterative planning framework outperforms both baseline methods across all three metrics. Our method achieves the highest diversity (0.898), exceeding NOVA (0.867) and AI-Researcher (0.680), which indicates a stronger ability to explore a wider range of research directions. It also attains the highest novelty (0.133), compared with 0.107 for NOVA and 0.067 for AI-Researcher, suggesting that the framework is more effective in generating ideas that differ from existing literature.

The advantage is even more pronounced in terms of high-score ratio. Our method reaches 0.184, substantially higher than NOVA (0.026) and AI-Researcher (0.013), indicating that the generated ideas are not only more diverse and novel but also more likely to be judged as high-quality research proposals.

Overall, these results demonstrate that multi-agent iterative planning and structured knowledge search can effectively improve both the exploratory breadth and the quality of LLM-generated research ideas.

### 4.4 Cross-Model Generalizability

To examine whether the effectiveness of the proposed framework depends on a specific backbone model, we further instantiate the same multi-agent iterative planning framework with three different LLMs: DeepSeek-3.1, GPT-4o (2024-11-20), and qwen3-8b. All models are evaluated under the same experimental setup, and the results are reported in Figure 4.

As shown in Figure 4, the proposed framework remains effective across different backbone models, suggesting that its utility does not rely solely on the capability of a single LLM. Among the three models, DeepSeek-3.1 achieves the highest Diversity score (0.898) and the highest HighScoreRatio (0.184), indicating a stronger ability to explore a broader range of research directions while also generating a larger proportion of high-quality ideas. In contrast, GPT-4o and qwen3-8b obtain higher Novelty scores, reaching 0.151 and 0.183, respectively, which suggests that different backbone models may exhibit different strengths when instantiated within the same iterative framework.

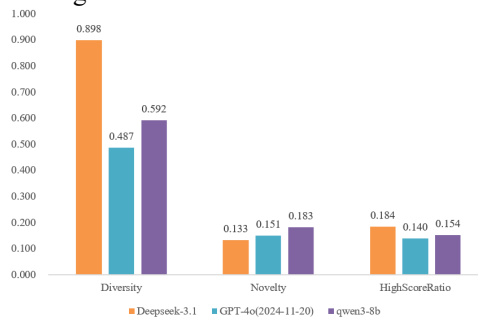

Figure 4. Performance of the proposed multi-agent iterative planning framework with different backbone models on three evaluation metrics: Diversity, Novelty, and HighScoreRatio.

Overall, these results provide preliminary evidence that the proposed framework generalizes across heterogeneous LLM backbones, including both a strong proprietary model and a smaller open-source model. Although the absolute performance varies across models, all three backbones are able to generate competitive research ideas under the same framework, indicating that the effectiveness of the method is not restricted to DeepSeek-3.1 alone.

### 4.5 Comparative validation with top conference paper ideas

To further evaluate the practical quality of the research ideas generated by the proposed framework, we selected 402 ideas from the final round of the multi-agent iterative planning process with Swiss-system scores of no less than 3 out of 5 and compared them with research ideas extracted from papers submitted to ICLR 2025. The threshold of 3 was adopted for three reasons. First, under the Swiss-system evaluation, ideas with scores of 3 or above have already demonstrated a certain level of competitiveness, and can therefore be regarded as medium- to high-quality outputs. Second, this threshold roughly corresponds to a baseline standard of academic acceptability, making the comparison more relevant to real conference evaluation scenarios. Third, removing clearly low-quality or immature ideas allows for a more accurate assessment of the framework's ability to generate research ideas with potential academic value.

Before comparing generated ideas with conference paper ideas, we first examined whether the evaluation strategy used in this study—namely, the combination of cross-group Swiss pairing and zero-shot LLM-based pairwise comparison—was capable of distinguishing accepted papers from rejected papers. Specifically, among the accepted NLP-related papers at ICLR 2025, we selected a total of 402 ideas, including all 61 oral papers, all 88 spotlight papers, and 253 randomly sampled poster papers. At the same time, 402 ideas were randomly sampled from the 731 rejected NLP-related submissions to ensure balanced sample sizes and comparability between groups. A five-round cross-group Swiss pairing procedure was then conducted, and the results are reported in Table 2.

**Table 2.** Comparison of Ideas in Accepted vs. Rejected Papers at ICLR 2025

| Group | Mean (S.D.) | T | Significance |
|---|---|---|---|
| Ideas for accepted papers | 3.256(1.095) | 57.27 | 0.000*** |
| Ideas for rejected papers | 1.744(1.148) | | |

Note: * indicates significance at the 0.05 level, ** indicates significance at the 0.01 level, *** indicates significance at the 0.001 level.

Based on this validation, we further conducted two comparative analyses. The first compared the generated ideas with ideas from accepted ICLR 2025 papers, and the second compared the generated ideas with ideas from rejected ICLR 2025 papers. As shown in Table 3, accepted paper ideas achieved a significantly higher mean score (2.776, SD = 1.646) than the ideas generated by our framework (2.224, SD = 0.777), and the difference was highly significant (t = 10.390, p < 0.001). This result indicates that, although the generated ideas have reached a reasonable quality level, they still fall short of the average quality of accepted papers at a top-tier conference. At the same time, the relatively large standard deviation among accepted paper ideas suggests greater variation in the quality distribution of accepted submissions.

As shown in Table 4, the generated ideas obtained a significantly higher mean score (2.689, SD = 0.818) than rejected paper ideas (2.311, SD = 1.620), with the difference again reaching a high level of statistical significance (t = 7.910, p < 0.001). This finding suggests that the ideas produced by the proposed framework are, on average, superior to those found in rejected conference submissions. In addition, the larger standard deviation observed among rejected paper ideas indicates that their quality distribution is more dispersed, with substantial variation across submissions.

Taken together, these results show that the quality of the ideas generated by our framework lies between that of accepted and rejected ICLR 2025 papers. Although the generated ideas have not yet reached the average level of accepted papers, they significantly outperform rejected papers. This suggests that the proposed framework is already capable of generating research ideas with a certain degree of academic competitiveness. Nevertheless, there remains room for improvement in terms of innovation depth, technical sophistication, and breakthrough potential when benchmarked against the acceptance standards of top-tier conferences.

**Table 3.** Comparison of paper ideas accepted at the 2025 ICLR conference with ideas generated in this study.

| Group | Mean (S.D.) | T | Significance |
| --- | --- | --- | --- |
| Ideas for accepted papers | 2.776(1.646) | 10.390 | 0.000*** |
| Generated ideas | 2.224(0.777) | | |

Note: * indicates significance at the 0.05 level, ** indicates significance at the 0.01 level, *** indicates significance at the 0.001 level.

**Table 4.** Comparison of paper ideas rejected at the 2025 ICLR conference with ideas generated in this study.

| Group | Mean (S.D.) | T | Significance |
| --- | --- | --- | --- |
| Generated ideas | 2.689(0.818) | 7.910 | 0.000*** |
| Ideas for rejected papers | 2.311(1.620) | | |

Note: * indicates significance at the 0.05 level, ** indicates significance at the 0.01 level, *** indicates significance at the 0.001 level.

### 4.6 Impact of Different Agent Team Sizes on Metrics

We examine the impact of varying agent team sizes on performance metrics by analyzing the best-performing third iteration results. As shown in Figure 5, for diversity, as the team size increases from 2 to 8, the uniqueness ratio exhibits an overall declining trend, starting from a relatively high level and gradually decreasing. This suggests that larger team sizes may lead to a reduction in uniqueness, which is likely related to the inherent knowledge limitations of LLMs. Generating more content increases the likelihood of similarity, indicating that expanding the scale of multi-agent systems does not necessarily enhance the uniqueness of LLM-generated content. This reflects a trade-off between quality and uniqueness.

For novelty, no clear trend is observed in relation to team size. However, the overall values remain relatively low and stable, indicating that team size has an insignificant impact on novelty. This further suggests that the proposed method cannot improve novelty by scaling up the number of agents.

The proportion of high-quality ideas fluctuates between 0.2 and 0.3 as the team size varies from 2 to 8, without showing a clear linear increase or decrease. However, in local variations:

Small teams (team size of 2-3): The proportion of high-quality ideas is relatively low, around 0.2. This may be due to limited resources and manpower in smaller teams, making it difficult to achieve high performance across all aspects.

Medium-sized teams (team size of 4-7): The proportion of high-quality ideas increases and stabilizes around 0.25. At this scale, teams may achieve a better balance in personnel allocation and collaboration, leading to improved overall performance.

Large teams (team size of 8): The proportion of high-quality ideas drops back to around 0.2. This may be attributed to increased management complexity and communication costs in larger teams, which can negatively impact overall efficiency and quality.

These findings align with the conclusion that an optimal team size can facilitate the generation of impactful research (L. Wu et al., 2019).

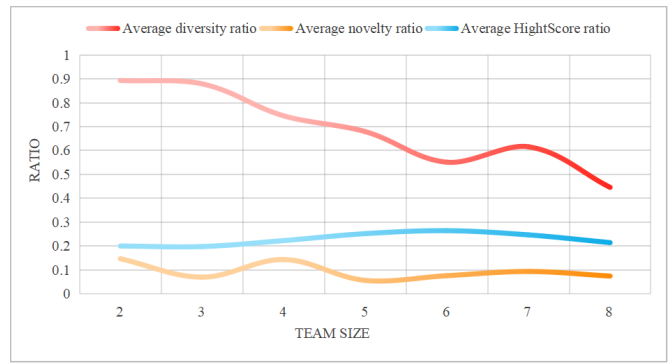

**Figure. 5.** Trend of Metrics Across Different Team Sizes

### 4.7 The Impact of Iteration Count on Metrics

As shown in Figure 6, the number of iterations has a significant impact on all metrics. The average diversity ratio peaks during the second iteration and then slightly declines. The average novelty ratio shows a notable improvement in the second iteration, with a marginal increase in the third iteration. Meanwhile, the proportion of high-quality ideas gradually rises with each iteration. These results suggest that the proposed method retains potential for generating high-quality ideas, though it exhibits some limitations in terms of novelty and diversity.

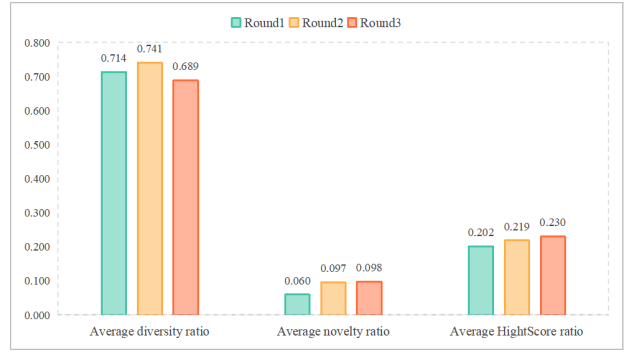

**Figure. 6.** Impact of Iteration Count on Average Metrics

Across different iteration counts, as illustrated in Figure 7, the best performance in diversity and novelty metrics consistently occurs in smaller teams, while the highest proportion of high-quality ideas is consistently achieved by teams of 5-7 members. This indicates that the multi-agent strategy holds promise for enhancing the quality of research ideas generated by LLMs.

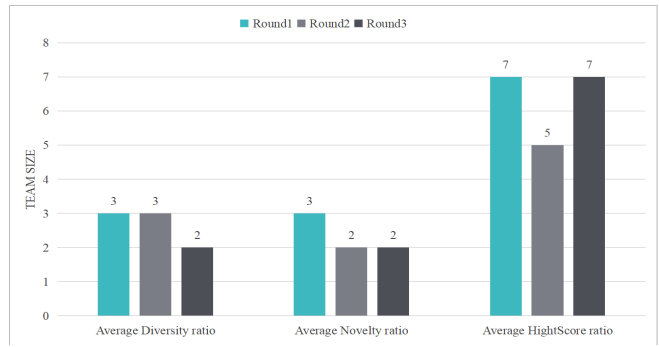

**Figure. 7.** Variation in Team Size Corresponding to the Best Metrics per Iteration

### 4.8 Ablation Study

We answer RQ2 in this section. The multi-agent iterative planning and search strategy integrates two core modules: knowledge planning and search, and multi-agent generation. A key objective of this study is to determine which module plays a decisive role in influencing critical metrics. To this end, we set the number of agents to 1, focusing on the impact of a single agent combined with knowledge planning and search on research idea generation. The best performance of the multi-agent iterative planning and search strategy is used as a benchmark for comparison.

As shown in Figures 8(a), 8(b), and 8(c), the single-agent approach outperforms in terms of diversity and novelty metrics. This suggests that the knowledge planning and search module positively enhances the generative capabilities of LLMs, indicating that combinatorial knowledge effectively guides LLMs. However, we also observe a declining trend in the performance of the single-agent system as the experiment progresses, suggesting that it may encounter bottlenecks in the research idea generation process. This finding aligns with the conclusions of Hu et al. (2024) in their study on Nova, where performance similarly plateaued after a certain number of iterations.

In contrast, while the multi-agent system slightly underperforms in diversity and novelty compared to the single-agent approach, it demonstrates significant advantages in the quality of generated ideas. Notably, the multi-agent system exhibits a consistent upward trend across all metrics. This indicates the potential of multi-agent systems and highlights the feasibility of incorporating innovative methodologies. In other words, combinatorial innovation theory and methodological approaches can effectively guide LLMs in the task of generating research ideas.

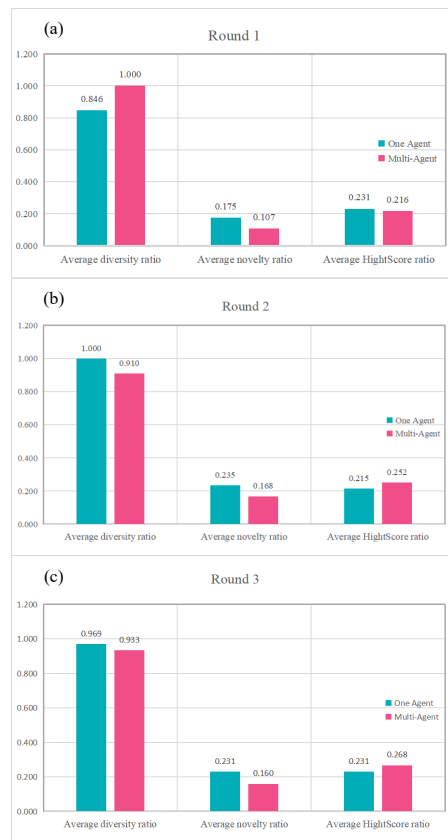

**Figure. 8.** Comparative Performance of Single-Agent vs. Multi-Agent Systems Across Iterations

## 4.9 Entity Recombination Analysis

Combinatorial innovation theory suggests that creative ideas rarely emerge from entirely new inventions; instead, they arise through atypical recombination of existing elements. In this study, entities such as methods, tasks, datasets, and metrics are treated as basic knowledge units, and their recombination reflects the core mechanism of combinatorial innovation. Therefore, tracing how entities are inherited, introduced, and reorganized across iterations provides a useful lens for understanding how the proposed framework generates innovative research ideas. Based on the comparative results in Section 4.4, we select three high-performing generated proposals for fine-grained entity-level analysis.

To support this analysis, we fine-tune the scientific entity extraction model Hgere (Yan et al., 2023) on the SciNLP dataset (Duan et al., 2025), which defines five entity types: Method, Task, Metric, Dataset, and Other. We then extract entities from the intermediate outputs of each iteration, including both the generated ideas and their corresponding knowledge bases. Here, the knowledge base consists of the previous-round idea together with the titles and abstracts of the newly retrieved literature. For each round, we calculate the entity overlap ratio between the generated idea and its knowledge base, and trace the source of each entity as originating from the last idea, retrieved references, or the LLM itself. Table 5 presents the entity counts of ideas and knowledge bases across iterations for the three selected cases.

**Table 5. Entity Counts of Ideas and Knowledge Bases Across Iteration Stages .**

| Idea_id | T1(idea) | T1(KB) | T2(idea) | T2(KB) | T3(idea) | T3(KB) |
|---|---|---|---|---|---|---|
| 1 | 51 | 487 | 58 | 561 | 60 | 335 |
| 2 | 48 | 297 | 56 | 422 | 61 | 946 |
| 3 | 54 | 289 | 61 | 338 | 58 | 410 |

Note: T1=Turn 1, T2=Turn 2, T3=Turn 3, KB= Knowledge Base.

To illustrate the recombination dynamics in detail, we present one representative case: **Causal Neuro-Symbolic Swarm Intelligence with Human-in-the-Loop Adaptation for Autonomous Disaster Response**. Table 6 summarizes the sources of entities in Idea 1 across the three turns. Figure 9 shows the entity structure of the initial idea.

**Table 6. The source of entities in Idea 1 .**

|  | Turn1 | Turn 2 | Turn 3 |
|---|---|---|---|
| Entity Counts of the Idea | 51 | 58 | 60 |
| From Last Idea | 6(11.8%) | 25(43.1%) | 47(78.3%) |
| From Reference | 9(17.6%) | 2(3.3%) | 3(5%) |
| From LLM | 36 | 31 | 10 |

Note: The percentage in parentheses indicates the entity overlap rate between the idea and its corresponding knowledge base.

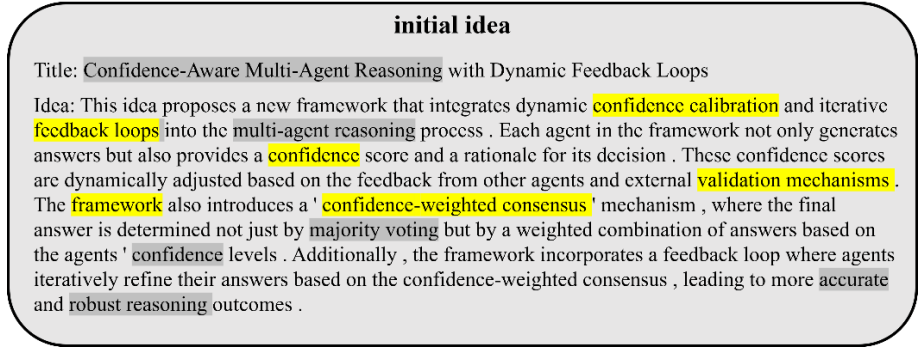

**Figure. 9.** The initial research idea of Idea 1. Note: Yellow blocks denote entities used by the Turn 1 idea, while gray blocks represent its own entities.

In Turn 1, the generated idea contains 51 entities, among which 11.8% come from the previous idea, 17.6% from retrieved references, and 70.6% are newly introduced by the LLM. This stage is characterized by broad exploratory recombination: while preserving the core logic of the initial idea, the model integrates external concepts such as neuro-symbolic reasoning, uncertainty quantification, and multi-agent systems, and further contributes a large number of self-generated entities related to architecture design, technical components, and evaluation criteria. Figure 10 visualizes the entity composition of the Turn 1 idea.

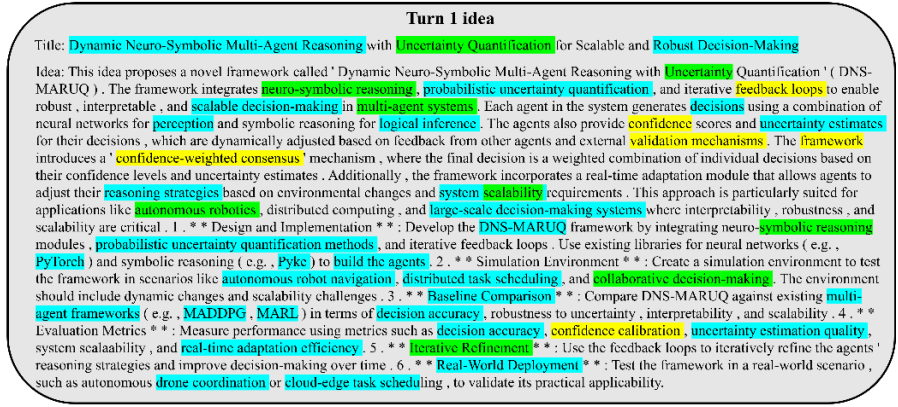

**Figure. 10.** The research idea from turn 1. Note: Blue blocks represent entities from the LLM, yellow blocks represent entities from the last idea, and green blocks represent entities from the reference literature. Stronger mechanisms to prevent excessive path dependence during iterative idea generation.

In Turn 2, the recombination pattern shifts substantially. The proportion of entities inherited from the previous round rises to 43.1%, while the share from references drops to 3.3%. This indicates that the model moves from broad exploratory generation toward a more focused refinement strategy. Rather than introducing many new theoretical concepts, the model consolidates the technical backbone established in Turn 1 and adapts it to a more concrete disaster-response setting by adding domain-specific entities such as drone swarms, environmental hazard monitoring, and scenario-oriented planning modules. At the same time, several technical refinements are introduced, including Bayesian neural networks and a domain-aware uncertainty propagation mechanism. Figure 11 illustrates the entity composition of the Turn 2 idea.

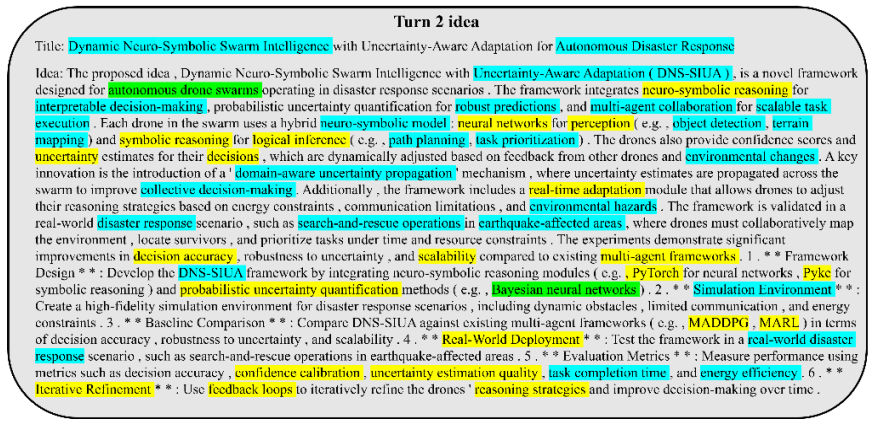

**Figure. 11.** The research idea from turn 2. Note: Blue blocks represent entities from the LLM, yellow blocks represent entities from the last idea, and green blocks represent entities from the reference literature.

In Turn 3, path dependence becomes even stronger. The proportion of inherited entities further increases to 78.3%, whereas only 5% of entities come from references and the number of newly generated entities sharply declines. Although the model still introduces several meaningful additions, such as causal reasoning, causal Bayesian networks, and human-in-the-loop interaction, the overall idea trajectory becomes highly locked into the previous technical pathway. This suggests that later iterations tend to emphasize deepening and extending an established line of reasoning rather than opening up new directions of recombination. Figure 11 shows the entity composition of the Turn 3 idea.

Overall, the case analysis reveals a clear three-stage evolutionary pattern in entity recombination. Turn 1 corresponds to exploratory recombination, where the model actively integrates cross-domain concepts and generates a large proportion of novel entities. Turn 2 reflects focused recombination, where the model preserves the core framework while adapting it to specific scenarios and adding contextualized technical details. Turn 3 exhibits path locking, in which excessive reliance on inherited entities constrains the introduction of substantially new components. This pattern is consistent with the results reported in Section 4.5: as iterations proceed, idea quality tends to improve, whereas diversity gradually declines.

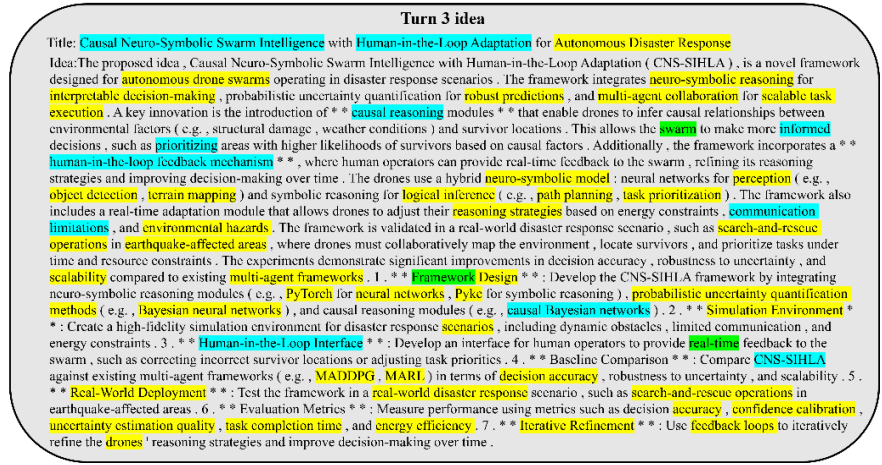

**Figure. 11.** The research idea from turn 3. Note: Blue blocks represent entities from the LLM, yellow blocks represent entities from the last idea, and green blocks represent entities from the reference literature.

These findings provide further insight into the mechanism of the proposed framework. Although iterative search and multi-agent refinement can effectively improve idea quality, the model does not always fully absorb newly retrieved external knowledge. Instead, it increasingly relies on inherited structures and internally generated extensions. This suggests that future work should focus on two directions: improving the effective assimilation of external knowledge by LLMs, and designing stronger mechanisms to prevent excessive path dependence during iterative idea generation.

### 4.10 Qualitative Error Analysis

To complement the quantitative evaluation reported above, we further examined a small set of generated ideas that did not meet the threshold for inclusion in the main comparative analysis. Our goal was not to identify extreme outliers, but to characterize representative failure modes that remain plausible under the current framework. The cases below suggest that, despite improvements in diversity and quality metrics, LLM-based research idea generation can still produce outputs that are technically inconsistent, insufficiently grounded in retrieved knowledge, or overly ambitious relative to their methodological support.

One recurring failure mode is logical inconsistency within the proposed method itself. For example, as shown in Appendix Figure. A17, one generated idea proposes using LLaMA-2-7B as the frozen base model while defining its routing mechanism in terms of an input-side [CLS] token embedding from the model's encoder. This formulation is internally inconsistent, because the named backbone does not naturally expose the encoder-style representation on which the rest of the mechanism depends. The issue is not merely missing implementation detail; rather, the central control component is built around an incompatible modeling assumption. This suggests that the system can preserve high-level architectural fluency while failing to maintain consistency between the selected backbone and the technical operations applied to it.

A second failure mode is hallucinated or weakly supported technical claims. In one temporally adaptive hypernetwork-based idea, as shown in Appendix Figure. A18, the proposal states that the model could make reasonable predictions for a future time period simply by feeding a future timestamp into the temporal controller. However, the idea does not define a forecasting objective, specify what type of future knowledge is assumed predictable, or explain why temporal conditioning alone should support extrapolation beyond observed data. The claim therefore extends well beyond what is justified by the retrieved literature and the proposed mechanism. This kind of error is particularly important because the output remains rhetorically convincing even when its technical warrant is limited.

A third failure mode concerns weak methodological coherence despite apparent novelty. One idea introduces a dynamic concept knowledge graph for continual learning, where latent activations are clustered into reusable "concepts," assigned graph structure, and then linked to replay examples through relevance-weighted distillation, as shown in Appendix Figure. A19. Although each component resembles a familiar research direction in isolation, the proposal leaves unclear how model-internal activations can be converted into stable concept units that are both interpretable and operationally useful for replay selection. In other words, the idea appears innovative at the component level, but the transitions between its components are under-specified. This suggests that the framework can generate methodologically rich descriptions without always securing a coherent bridge between representation, supervision, and evaluation.

A fourth failure mode is superficial novelty accompanied by low practical feasibility. For instance, as shown in Appendix Figure. A20, one generated idea for industrial health monitoring combines temporal expert routing, tensorized adapters, physics-informed priors, memory replay, multi-task continual learning, interpretability analysis, and edge-deployment constraints within a single proposal. While the combination appears highly novel, its scope is so broad that the contribution risks becoming diffuse, with multiple difficult subproblems bundled together before any one mechanism is validated. The resulting idea is not obviously impossible, but it lacks the degree of methodological focus typically expected of a credible research project. This case aligns with the broader tendency, also reflected in our turn-level case study, for later-stage ideas to accumulate additional technical machinery rather than sharpen a single testable contribution.

Taken together, these cases indicate that current LLM-based research idea generation systems may fail not only through low novelty, but also through more subtle forms of technical weakness, including internal inconsistency, unsupported extrapolation, underspecified component integration, and overextended design scope. These observations are consistent with the limitations discussed in Section 6.2: the present evaluation framework emphasizes novelty, diversity, and judged quality, but does not fully capture feasibility, methodological soundness, or the reliability of technical grounding. A qualitative error analysis therefore provides an important complement to metric-based evaluation by clarifying which aspects of failure remain difficult to detect under existing automatic or semi-automatic assessment procedures.

## 5 Demonstration

We develop an interactive demonstration system based on the proposed multi-agent iterative search framework for research idea generation. In the Hugging Face demo, the collaborative team size is automatically determined by the number of authors of the user-selected paper. The complete agent system consists of these team agents, together with one search agent and one evaluation agent. The maximum number of iterative refinement rounds is set to three, which clarifies the exact demo configuration and improves reproducibility.

As illustrated in Figure 12, the demonstration workflow begins with a user-provided research topic. The topic is first passed to an LLM-based literature search agent with autonomous search-path planning capability. By integrating the Semantic Scholar API, the agent performs topic-oriented literature retrieval and relevance verification, and then returns a refined list of highly relevant papers. The user selects one paper from this list as the basis for subsequent analysis. The system then applies the proposed multi-agent iterative search framework to analyze the selected paper and generate a novel research idea.

The interface adopts a two-column layout. The left panel displays the retrieved literature and allows users to select papers for further exploration, while the right panel provides the user–LLM interaction interface and presents the full reasoning and generation process in real time (see Appendix Figure A11).

To illustrate the workflow, we use Research Idea Generation as the input topic. The right panel shows the complete action trajectory of the literature search agent, including search planning, Semantic Scholar retrieval, and relevance verification. At the same time, the left panel presents the retrieved papers and their detailed information for user selection (Appendix Figure A12). In our example, we select "ResearchAgent": Iterative Research Idea Generation over Scientific Literature with Large Language Models as the input paper. The system then automatically extracts its abstract, references, and author information, generates initial research ideas, and synchronously displays the reasoning process. These initial ideas are further used to support deeper related-knowledge retrieval (Appendix Figure A13).

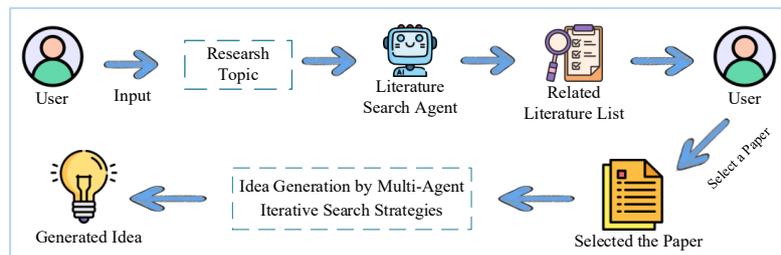

**Figure. 12.** Demonstration Process

Next, the initial research ideas and retrieved knowledge are provided to the virtual scientist agents, each of whom is instantiated with the background information of the selected paper's author(s). Based on both prior background knowledge and newly retrieved evidence, each agent independently proposes a research idea while exposing its reasoning process to the user (Appendix Figure A14). Finally, the generated ideas are compared through a competitive evaluation process to determine the final output idea (Appendix Figure A15). The full demo is publicly available on Hugging Face: https://huggingface.co/spaces/cshuai20/MAGenIdeas.

# 6 Discussion

### 6.1 Research Implications

**Theoretical Insights**: This study contributes to the emerging research on automated scientific ideation by introducing a framework that integrates combinatorial innovation theory with multi-agent reasoning. Specifically, the proposed approach models research idea generation as a process of structured knowledge recombination guided by iterative planning and collaborative evaluation.

The empirical results demonstrate that combining cross-domain knowledge planning with multi-agent collaboration significantly improves the diversity, novelty, and quality of generated research ideas compared with baseline systems. The ablation analysis further confirms that combinatorial knowledge planning expands the exploration space of LLMs, while multi-agent interaction mechanisms support iterative refinement and critical evaluation of intermediate ideas.

These findings provide empirical evidence that structured knowledge recombination and simulated collaborative reasoning can enhance the creative capabilities of LLM-based systems. More broadly, the results suggest that theoretical frameworks from innovation studies can play an important role in guiding the design of generative AI systems for scientific discovery.

**Practical Implications**: From a practical perspective, the results indicate that collaborative multi-agent architectures provide an effective strategy for improving LLM-based research idea generation systems. Compared with single-agent prompting approaches, the introduction of multiple agents with differentiated roles enables the system to simulate diverse cognitive perspectives during the ideation process. This mechanism promotes richer knowledge recombination and reduces the risk of prompt-induced conceptual bias.

Furthermore, grounding system design in established theoretical frameworks enhances both the interpretability and robustness of the generation process. Based on the proposed framework, we have also developed a prototype research idea generation system designed for human–computer interaction scenarios, demonstrating the potential of the approach to support researchers in early-stage scientific exploration.

### 6.2 Limitations and Future Research

Despite the promising results, this study has several limitations that open directions for future research.

**Domain Generalizability.** This study evaluates the proposed framework within the natural language processing (NLP) domain, which provides a well-controlled environment for studying research idea generation. However, different scientific disciplines vary significantly in their epistemic structures, knowledge organization patterns, and research methodologies. For instance, some domains rely more heavily on experimental protocols, specialized datasets, or domain-specific tools rather than primarily textual conceptual reasoning. These differences may affect the effectiveness of knowledge retrieval and recombination mechanisms in the proposed framework. Future research will extend the framework to additional scientific domains, such as computer vision, life sciences, and physics, and explore domain-adaptive strategies for knowledge integration.

**Reproducibility and Data Dependency.** Reproducing this research in other contexts may present challenges related to data availability and metadata completeness. The proposed framework relies on structured bibliographic information including references, citation relationships, and author profiles to construct the knowledge context used for idea generation. However, the coverage and consistency of such metadata may vary across different scholarly databases. In addition, variations in database APIs, citation coverage, and data preprocessing procedures may influence experimental reproducibility. Future studies may address these challenges by developing standardized data processing pipelines, releasing benchmark datasets for research idea generation, and designing database-agnostic retrieval modules capable of integrating heterogeneous scholarly data sources.

**Evaluation Scope.** Although this study evaluates generated research ideas using metrics of novelty, diversity, and quality, these indicators do not fully capture several important dimensions of scientific creativity and methodological soundness. In particular, they may fail to detect technical inconsistency, weakly supported claims, under-specified component integration, or low practical feasibility in otherwise fluent and apparently novel outputs. As shown in our qualitative error analysis, some generated ideas remain rhetorically convincing while lacking sufficient grounding or internal coherence. Future work may therefore incorporate expert-based review protocols, feasibility-oriented evaluation criteria, and stronger checks on technical consistency in order to provide a more comprehensive assessment framework for LLM-generated research ideas.

**Efficiency and latency trade-off**. Although the proposed framework improves idea quality through iterative retrieval, multi-agent refinement, and tournament-based evaluation, these gains come with additional computational overhead. As shown in Appendix Table. A16, stage-wise latency measurements show that runtime is mainly dominated by literature retrieval and multi-stage LLM inference. Under a 3-agent, 2-round setting, the full pipeline requires about 255-262 seconds after target paper selection, while a reduced setting without refinement and tournament still requires about 115 seconds. Parallelization substantially reduces the refinement-stage latency, but the end-to-end speedup remains limited because retrieval and evaluation still account for a large share of total runtime. Therefore, the current system should be viewed as a quality-oriented research assistant rather than a strictly real-time interactive system. In future work, this limitation may be mitigated through caching intermediate results, batching retrieval, adaptive agent budgets, early stopping, and broader parallel execution.

# 7 Conclusion

This study advances automated research idea generation by introducing a multi-agent iterative planning and search framework grounded in combinatorial innovation theory. Rather than treating idea generation as a one-shot prompting task, the proposed approach conceptualizes scientific ideation as a process of structured knowledge recombination, iterative retrieval, and collaborative evaluation across heterogeneous perspectives. In doing so, the study connects a theory of innovation with the design of LLM-based systems for scientific discovery.

The empirical findings show that the proposed framework consistently improves the diversity, novelty, and overall quality of generated research ideas relative to existing baselines. The results further suggest that combining cross-domain knowledge planning with differentiated agent perspectives helps expand the exploration space of LLMs and mitigates the limitations of single-perspective prompting. In addition, the comparison with accepted and rejected ICLR 2025 papers indicates that the generated ideas possess meaningful academic potential, even if they do not yet reach the level of the strongest accepted work.

Beyond performance gains, this study offers two broader contributions. First, it demonstrates that combinatorial innovation provides a useful theoretical lens for understanding how LLMs may support scientific creativity through knowledge recombination. Second, it shows that collaborative multi-agent architectures can serve as a practical design strategy for AI-assisted ideation systems by enabling iterative critique, refinement, and perspective diversification. These findings contribute to the growing literature on AI for science by showing that theory-informed system design can improve both the interpretability and the creative capacity of generative models.

Several limitations should also be acknowledged. The current evaluation is confined to the NLP domain, where textual reasoning and standardized publication conventions make the task comparatively tractable. In addition, the assessment of generated ideas focuses primarily on novelty, diversity, and quality, without fully capturing feasibility, practical value, or long-term scientific impact. The framework also depends on relatively complete bibliographic metadata and incurs nontrivial computational overhead due to iterative retrieval and multi-agent inference. Future work should therefore examine the transferability of the framework across disciplines, develop richer and more rigorous evaluation protocols, improve reproducibility under heterogeneous scholarly data conditions, and explore more efficient architectures for real-world deployment.

Overall, this study provides evidence that research idea generation can be meaningfully improved when LLM systems are guided not only by external knowledge, but also by principled mechanisms for knowledge recombination and simulated collaboration. We hope this work offers a useful foundation for future research on AI-assisted scientific ideation and broader forms of human–AI co-creation in knowledge-intensive domains.

## Ethical Statement

All literature data and author background information used in this study were sourced from publicly available academic databases, and none of the content involves personal privacy or sensitive information. During data processing, we strictly adhered to the terms of use and academic ethics guidelines of each database to ensure no risk of privacy breaches. To protect scholars' personal information security, all author names were anonymized during the analysis.

It is important to emphasize that the system developed in this study is solely intended to assist scientific research. Its design purpose is to provide research support for scholars, not to replace human researchers. Throughout its operation, the system emphasizes the importance of human oversight mechanisms, ensuring the quality of research results through human-machine collaboration.

## Reproducibility and Data Provenance

All data used in this study come from publicly accessible academic databases, including ACL Anthology, OpenAlex, Semantic Scholar, and OpenReview; no proprietary or restricted-access resources are involved. The data collection pipeline is deterministic and reproducible. Target papers are identified from the ACL 2024 long paper track, and bibliographic metadata, references, citation counts, and author profiles are enriched through OpenAlex and Semantic Scholar using paper titles and DOIs as identifiers. ICLR 2025 publication decisions and review outcomes are obtained from publicly available OpenReview fields. To protect privacy, all author names are anonymized in the released dataset, and no sensitive information beyond public academic profiles is collected. The code repository includes the full data processing scripts, filtering criteria, and domain-classification prompts to support reproducibility. We also considered the potential impact of retracted or discredited publications. Although no retracted papers were found in the current dataset, future versions of the framework could incorporate explicit filtering based on publication status or external retraction databases to further strengthen source integrity.

## Acknowledgments

This paper was supported by the National Natural Science Foundation of China (Grant No.72074113) and Open Foundation of the ISTIC-CLARIVATE Analytics Joint Laboratory for Scientometrics (No. IC2502). This paper is an extended version of the ISSI 2025 paper "*Enhancing Research Idea Generation through Combinatorial Innovation and Multi-Agent Iterative Search Strategies*. In: Proceedings of the 20th International Conference on Scientometrics and Informetrics (ISSI 2025), Yerevan, Armenia, 2025."

# References

Ajith, A., Xia, M., Chevalier, A., Goyal, T., Chen, D., & Gao, T. (2024, November). *LitSearch: A Retrieval Benchmark for Scientific Literature Search*. In Proceedings of the 2024 Conference on Empirical Methods in Natural Language Processing, Miami, Florida, USA.

Altshuller. (1984). *Creativity As an Exact Science*. In.

Baek, J., Jauhar, S. K., Cucerzan, S., & Hwang, S. J. (2025, April). *ResearchAgent: Iterative Research Idea Generation over Scientific Literature with Large Language Models*. In Proceedings of the 2025 Conference of the Nations of the Americas Chapter of the Association for Computational Linguistics: Human Language Technologies (Volume 1: Long Papers), Albuquerque, New Mexico.

Chan, C.-M., Chen, W., Su, Y., Yu, J., Xue, W., Zhang, S., Fu, J., & Liu, Z. J. a. p. a. (2023). *Chateval: Towards better llm-based evaluators through multi-agent debate*. In Proceedings of the twelfth International Conference on Learning Representations.

Christensen, C. M. (1997). *The Innovator's Dilemma*. In.

Duan, D., Peng, J., Zhang, Y., & Zhang, C. (2025). *SciNLP: A Domain-Specific Benchmark for Full-Text Scientific Entity and Relation Extraction in NLP*. In.

Funk, R. J., & Owen-Smith, J. J. M. S. (2017). A Dynamic Network Measure of Technological Change. *63*, 791-817.

Gu, X., & Krenn, M. (2024). Generation and human-expert evaluation of interesting research ideas using knowledge graphs and large language models. *arXiv preprint arXiv:.17044*.

Guo, S., Shariatmadari, A. H., Xiong, G., Huang, A., Kim, M., Williams, C. M., Bekiranov, S., & Zhang, A. (2025). *Ideabench: Benchmarking large language models for research idea generation*. In Proceedings of the 31st ACM SIGKDD Conference on Knowledge Discovery and Data Mining, Toronto on Canada.

Hu, X., Fu, H., Wang, J., Wang, Y., Li, Z., Xu, R., Lu, Y., Jin, Y., Pan, L., & Lan, Z. (2025, July). *NOVA: An Iterative Planning Framework for Enhancing Scientific Innovation with Large Language Models*. In Findings of the Association for Computational Linguistics: ACL 2025, Vienna, Austria.

Ke, Q., Pan, T., & Mao, J. J. R. P. (2026). The geography of novel and atypical research. *55*(1), 105345.

Kojima, T., Gu, S. S., Reid, M., Matsuo, Y., & Iwasawa, Y. (2022). Large language models are zero-shot reasoners. *Advances in neural information processing systems, 35*, 22199-22213.

Larivière, V., Archambault, É., Gingras, Y., & technology. (2008). Long-term variations in the aging of scientific literature: From exponential growth to steady-state science (1900–2004). *Journal of the American Society for Information Science, 59*(2), 288-296.

Lee, Y.-N., Walsh, J. P., & Wang, J. J. R. P. (2015). Creativity in scientific teams: : Unpacking novelty and impact. *44*, 684-697.

Li, R., Jing, L., Han, C., Zhou, J., & Du, X. (2024). Learning to generate research idea with dynamic control. *arXiv preprint arXiv:.14626*.

Liusie, A., Manakul, P., & Gales, M. (2023). Zero-shot NLG evaluation through pairware comparisons with llms. . *arXiv preprint arXiv:.07889*.

Lu, C., Lu, C., Lange, R. T., Foerster, J., Clune, J., & Ha, D. (2024). The ai scientist: Towards fully automated open-ended scientific discovery. *arXiv preprint arXiv:.06292*.

Prabhakaran, T., Lathabai, H. H., Changat, M. J. T. F., & Change, S. (2015). Detection of paradigm shifts and emerging fields using scientific network: A case study of Information Technology for Engineering. *91*, 124-145.

Prabhakaran, T., Lathabai, H. H., George, S., & Changat, M. J. S. (2018). Towards prediction of paradigm shifts from scientific literature. *117*, 1611 - 1644.

Pu, K., Feng, K. K., Grossman, T., Hope, T., Dalvi Mishra, B., Latzke, M., Bragg, J., Chang, J. C., & Siangliulue, P. (2025). *Ideasynth: Iterative research idea development through evolving and composing idea facets with literature-grounded feedback*. In Proceedings of the 2025 CHI Conference on Human Factors in Computing Systems.

Sadler-Smith, E. J. C. r. j. (2015). Wallas' four-stage model of the creative process: More than meets the eye? *Creativity research journal, 27*(4), 342-352.

Schmidgall, S., Su, Y., Wang, Z., Sun, X., Wu, J., Yu, X., Liu, J., Moor, M., Liu, Z., & Barsoum, E. (2025). Agent laboratory: Using llm agents as research assistants. *arXiv preprint arXiv:.04227*.

Schumpeter, J. A. J. A. e. (1964). Business Cycles: A theoretical, historical and statistical analysis of the Capitalist process, 1939. *4*.

Si, C., Yang, D., & Hashimoto, T. (2024). *Can llms generate novel research ideas? a large-scale human study with 100+ nlp researchers*. In Proceedings of the Thirteenth International Conference on Learning Representations, Singapore EXPO.

Su, H., Chen, R., Tang, S., Yin, Z., Zheng, X., Li, J., Qi, B., Wu, Q., Li, H., Ouyang, W., Torr, P., Zhou, B., & Dong, N. (2025, July). *Many Heads Are Better Than One: Improved Scientific Idea Generation by A LLM-Based Multi-Agent System*. In Proceedings of the 63rd Annual Meeting of the Association for Computational Linguistics (Volume 1: Long Papers), Vienna, Austria.

Uzzi, B., Mukherjee, S., Stringer, M., & Jones, B. J. S. (2013). Atypical combinations and scientific impact. *342*(6157), 468-472.

Wang, P., Li, L., Chen, L., Cai, Z., Zhu, D., Lin, B., Cao, Y., Liu, Q., Liu, T., & Sui, Z. (2023). Large language models are not fair evaluators. *arXiv preprint arXiv:.17926*.

Wang, P., Wang, X., Lou, C., Mao, S., Xie, P., & Jiang, Y. (2024, November). *Effective Demonstration Annotation for In-Context Learning via Language Model-Based Determinantal Point Process*. In Proceedings of the 2024 Conference on Empirical Methods in Natural Language Processing, Miami, Florida, USA.

Wang, X., Wei, J., Schuurmans, D., Le, Q., Chi, E., Narang, S., Chowdhery, A., & Zhou, D. (2022). *Self-consistency improves chain of thought reasoning in language models*. In Proceedings of the eleventh International Conference on Learning Representations.

Wei, J., Wang, X., Schuurmans, D., Bosma, M., Xia, F., Chi, E., Le, Q. V., & Zhou, D. (2022). Chain-of-thought prompting elicits reasoning in large language models. *Advances in neural information processing systems, 35*, 24824-24837.

Wu, L., Wang, D., & Evans, J. A. (2019). Large teams develop and small teams disrupt science and technology. *Nature, 566*(7744), 378-382.

Wu, N., Gong, M., Shou, L., Liang, S., & Jiang, D. (2023). *Large language models are diverse role-players for summarization evaluation*. In Proceeding of international conference on natural language processing and Chinese computing.

Yan, Z., Yang, S., Liu, W., & Tu, K. J. a. p. a. (2023). Joint entity and relation extraction with span pruning and hypergraph neural networks.

Yao, S., Yu, D., Zhao, J., Shafran, I., Griffiths, T., Cao, Y., & Narasimhan, K. (2023). Tree of thoughts: Deliberate problem solving with large language models. *Advances in neural information processing systems, 36*, 11809-11822.

Yuan, J., Yan, X., Zhang, B., Chen, T., Shi, B., Ouyang, W., Qiao, Y., Bai, L., & Zhou, B. (2025, July). *Dolphin: Moving Towards Closed-loop Auto-research through Thinking, Practice, and Feedback*. In Proceedings of the 63rd Annual Meeting of the Association for Computational Linguistics (Volume 1: Long Papers), Vienna, Austria.

Zhang, Z., Zhang, A., Li, M., & Smola, A. (2022). Automatic chain of thought prompting in large language models. arXiv 2022. *arXiv preprint arXiv:.03493*.

Zhou, D., Schärli, N., Hou, L., Wei, J., Scales, N., Wang, X., Schuurmans, D., Cui, C., Bousquet, O., & Le, Q. (2022). *Least-to-most prompting enables complex reasoning in large language models*. In Proceedings of the eleventh International Conference on Learning Representations.

Zhu, K., Wang, J., Zhao, Q., Xu, R., & Xie, X. (2024). *Dynamic evaluation of large language models by meta probing agents*. In Proceeding of the Forty-first International Conference on Machine Learning.

# Appendix

> System prompt: You are an expert researcher in AI. Your goal is to propose some innovative and valuable research ideas based on the target paper.
>
> Follow these steps to generate innovative research ideas for exploration.
> Understand the Target Paper and Related Works. Target Paper: This is the core research study you aim to enhance or build upon. It serves as the foundation for identifying and developing new research ideas. Referenced Papers: These are studies cited by the target paper, providing additional context and insights directly relevant to the primary research topic. They are crucial for understanding and expanding upon the target paper.
> Leverage Scientific Discovery Theories. Select appropriate scientific discovery theories and combine them with insights from the target paper to generate creative, impactful, and feasible research ideas.
> Explore Scientific Discovery Methodologies. Below are 10 general laws and methodologies of scientific discovery from the philosophy of science. Choose one or more of these methodologies to propose new research ideas for the target paper.
> Select and Propose New Ideas. Identify the 5 most suitable theories or methods for the target paper and propose 15 new research ideas based on them.
> Requirements:
> Output: Provide 15 new research ideas worth exploring.
> Quality: Aim for ideas with the potential to win best paper awards at top conferences like ACL, NeurIPS, ICLR, and CVPR.
> Relevance: Skip theories that do not align well with the target paper. Ensure the chosen theories and methods are logical and appropriate for the research context.
> Input:
> Target Paper: {target_paper}
> References: {references}
> Scientific Discovery Theory: {scientific_discovery_theory}
> Response Format:
> Thought: <THOUGHT>
> IDEA: ```json<JSON>```
> In <THOUGHT>, explain your thought process, including why you chose specific theories and why the proposed ideas have the potential to win best paper awards at top conferences.
> In <JSON>, provide the new idea with the following fields:
> Title: A concise title for the idea, suitable for a research paper.
> Idea: A detailed description of the idea, including its significance and potential impact.
> Thinking: A detailed explanation of the thought process, including the scientific discovery theory used.
> Rationale: A detailed justification for why this idea is innovative and feasible.
> Ensure the JSON format is precise, as it will be automatically parsed.

**Figure. A1.** Prompt for initial Research Idea Generation

| |
|---|
| **Title**: Beyond Linguistic Features: Leveraging Cognitive Biases for Machine-Generated Text Detection |
| **Idea**: This idea proposes a novel approach to detecting machine-generated text by leveraging cognitive biases inherent in human language processing. While current methods focus on linguistic features, this approach hypothesizes that machine-generated text lacks subtle cognitive biases that humans exhibit, such as confirmation bias, anchoring, and availability heuristic. By training models to detect these biases, we can create a more robust detector that performs well even when linguistic distinctions are minimal. The idea involves collecting a dataset of human and machine-generated texts annotated for cognitive biases, developing a model to detect these biases, and evaluating its performance in diverse scenarios. |
| **Thinking**: This idea is inspired by Pierce\u2019s hypothetical deduction method and Kuhn\u2019s paradigm theory. The hypothesis is that machine-generated text lacks cognitive biases, which are subtle but consistent features of human language. By shifting the paradigm from linguistic features to cognitive biases, we can address the anomaly of decreasing linguistic distinctions. The rationale is that while machines can mimic linguistic patterns, they cannot replicate the nuanced cognitive biases that humans naturally exhibit. |
| **Rationale**: Current detection methods struggle with out-of-distribution texts and are limited by the decreasing linguistic distinctions between human and machine-generated text. By focusing on cognitive biases, which are deeply ingrained in human language, we can create a more robust detection method that is less reliant on surface-level linguistic features. This approach has the potential to significantly improve detection accuracy in real-world scenarios, making it a strong candidate for a best paper award. |

**Figure. A2.** An example for initial research idea generation

| |
|---|
| System prompt: You are now a senior scholar in the field of artificial intelligence, and you need to develop a search strategy based on a given research idea |
| Task: To analyze a given research idea and determine which fields of papers should be retrieved to collect comprehensive information and new knowledge. This will provide a basis for further research and the discovery of new ideas.<br>Input:<br>Research Idea: {idea}<br>Response Format:<br>Thought: <THOUGHT><br>Search Plans: ```json<JSON> ```<br>In <THOUGHT>, explain the reasoning behind the search strategy, including why specific fields and keywords were chosen.<br>In <JSON>, provide a detailed search plan with the following fields:<br>Search Plan: A brief description of the overall search strategy.<br>Search Fields: A list of fields relevant to the research idea.<br>Search Keywords: A structured list of keywords for each field, formatted as:<br>"Search_Keywords": [{"Field": "field_name", "Keywords": ["keyword1", "keyword2", "keyword3"]}]<br>Ensure the JSON format is precise, as it will be automatically parsed. |

**Figure. A3.** Prompt for knowledge planning and search

> Search Plan: "To gather comprehensive information, we will search for papers in cognitive biases, machine-generated text detection, NLP, dataset creation, and model evaluation. Each field will be explored using specific keywords to ensure relevance and depth."
>
> Search Keyword": [{"Field": "Cognitive Biases in Human Language Processing", "Keywords": ["confirmation bias", "anchoring bias", "availability heuristic", "cognitive biases in language", "human language processing"]},{"Field": "Machine-Generated Text Detection", "Keywords": ["machine-generated text", "AI-generated text", "text detection", "linguistic features", "deepfake text"]},{"Field": "Natural Language Processing (NLP)","Keywords": ["text analysis", "language models", "NLP techniques", "text classification", "semantic analysis"]}.
>
> Thought: The research idea revolves around detecting machine-generated text by leveraging cognitive biases in human language processing. To develop a comprehensive search strategy, we need to explore several interconnected fields: (1) **Cognitive Biases in Human Language Processing** to understand the theoretical foundation of the biases, (2) **Machine-Generated Text Detection** to review existing methods and their limitations, (3) **Natural Language Processing (NLP)** for techniques to analyze and model text, (4) **Dataset Creation and Annotation** for methodologies to collect and label data, and (5) **Model Evaluation** for metrics and frameworks to assess detector performance. This strategy ensures a holistic understanding of the problem and provides a foundation for developing a novel approach.

**Figure. A4.** An example for knowledge planning and search

> System prompt: Your name is Scientist0, you belong to following affiliations ['Westlake University'], you have researched on following topics ['Natural Language Processing Techniques', 'Topic Modeling', 'Multimodal Machine Learning Applications', 'Text Readability and Simplification', 'Speech and dialogue systems' ], you have published 24 papers, you have 844 citations.
>
> Task: You are an ambitious scientist who is looking to propose a new idea that will contribute significantly to the field. Improve the seed idea or come up with the next impactful and creative idea for publishing a paper that will contribute significantly to the field by integrating your own knowledge and insights with the information provided.
> Here is the seed idea: {seed_idea}
> Here are the bad reviews of the seed idea, please improve according to the reviews: {bad reviews}
> You may refer to the following listed references to design a new idea or concept. These references can serve as inspiration, but you are not allowed to directly copy or replicate their content. Ensure that your design is original and addresses a specific problem or meets a unique need. References: {references}
> Please respond in the following format:
> Thought: <THOUGHT>
> New Idea: ```json<JSON>```
> In <THOUGHT>, briefly discuss your intuitions and motivations for the idea. Justify how this idea differs from existing ones, highlighting its unique aspects.
> In <JSON>, provide the new idea with the following fields and provide as many details as possible:
> Idea: A detailed description of the idea, outlining its significance and potential impact.
> Title: A title for the idea, will be used for the paper writing.
> Experiment: An outline of the implementation process. Describe your high-level design plan, including necessary design steps and the ideal outcomes of the experiments.
> Excitement: {excitement scoring rules} - "Excitement Rationale": {prompt for rationale}
> Feasibility: {feasibility scoring rules} - "Feasibility Rationale": {prompt for rationale}
> Novelty: {novelty scoring rules} - "Novelty Rationale": {prompt for rationale}
> Be cautious and realistic on your ratings. This JSON will be automatically parsed, so ensure the format is precise, and the content should be longer than 600 words. You only need to output one idea.

**Figure. A5.** Prompt for research idea generation

| |
|---|
| Title: Multimodal Few-Shot Learning for Cognitive Bias-Based Detection of Machine-Generated Text |
| Idea: This idea proposes a multimodal, few-shot learning framework for detecting machine-generated text by leveraging cognitive biases and contextual cues across text, speech, and visual modalities. Unlike existing methods that focus solely on linguistic features, this approach integrates psychological patterns (e.g., confirmation bias, anchoring) with multimodal data to create a more robust and adaptive detection system. The framework involves: (1) collecting a diverse dataset of human and machine-generated content annotated for cognitive biases and contextual features, (2) developing a multimodal deep learning model that combines text, speech, and visual embeddings, and (3) employing few-shot learning techniques to enable the model to generalize across new domains and languages. The system will be evaluated in real-time scenarios, such as social media moderation and customer service chatbots, to assess its effectiveness in distinguishing machine-generated content from human-authored text. This approach has the potential to significantly improve the accuracy and adaptability of machine-generated text detection, particularly in dynamic and multilingual environments. |
| Experiment: 1. Dataset Collection: Curate a multimodal dataset containing text, speech, and visual data from both human and machine-generated sources. Annotate the dataset for cognitive biases and contextual features. 2. Model Development: Design a multimodal deep learning model that integrates text, speech, and visual embeddings using transformer architectures. Incorporate few-shot learning techniques to enable domain adaptation. 3. Training and Evaluation: Train the model on the annotated dataset and evaluate its performance in detecting machine-generated text across diverse scenarios, including social media moderation and customer service chatbots. 4. Real-Time Testing: Deploy the model in real-time applications to assess its effectiveness and adaptability. Measure accuracy, recall, and F1-score as key performance metrics. 5. Comparative Analysis: Compare the proposed model with state-of-the-art methods to demonstrate its superiority in terms of accuracy and adaptability. |
| Excitement: 9, <br> Excitement Rationale: "This idea is highly exciting because it addresses a critical challenge in NLP and multimodal learning by integrating cognitive biases with advanced machine learning techniques. It has the potential to significantly improve the detection of machine-generated text in real-time applications, which is increasingly important in the era of large language models and AI-generated content. The combination of multimodal data and few-shot learning makes this approach highly adaptable and generalizable, setting it apart from existing methods.", |
| Feasibility: 7, <br> Feasibility Rationale: "The idea is feasible but requires careful planning and efficient use of resources. While the multimodal dataset collection and model development are achievable within 1-2 months, the real-time testing and deployment may require additional computational resources. The use of APIs and pre-trained models can help mitigate some of these challenges, but the integration of few-shot learning techniques may require advanced expertise." |

**Figure. A6.** An example for research idea generation

> System prompt: You are a reviewer specialized in Natural Language Processing. You are given two project summaries. One of them is accepted by a top AI conference (like ICLR or ACL) and the other one is rejected. Your task is to identify the one that has been accepted.
> Now decide which one is the accepted idea. give a meta-review to each paper.
> ```json{
> Decision: <DECISION>
> ReviewForPaper1: <ReviewForPaper1>
> ReviewForPaper2: <ReviewForPaper2> } ```
> In <DECISION>, return a number 1 or 2 and nothing else
> In <ReviewForPaper1>, write the review for paper 1, reasons for acceptance or rejection.
> In <ReviewForPaper2>, write the review for paper 2, reasons for acceptance or rejection.
> This JSON will be automatically parsed, so ensure the format is precise.

**Figure. A7.** Prompt for comparison of research ideas

> System prompt: You are now a senior scholar in the field of artificial intelligence.
> You are an ambitious scientist who will generate a summary based on given research idea and experimental steps.
> Requirements: The content of the abstract should cover: research questions and objectives, research methods, expected research results, and conclusions. Do not exceed 300 words.
> Here is the research idea: {idea};
> Here is the experimental steps: {experiment};
> Please respond in the following format:
> Thought: <THOUGHT>
> Abstract: ```json<JSON>```
> In <THOUGHT>, please briefly describe your thinking.
> In <JSON>, provide the abstract with the following fields:
> Title: A title for the abstract.
> Abstract: abstract.
> Be cautious and realistic on your ratings. This JSON will be automatically parsed, so ensure the format is precise.

**Figure. A8.** Prompt for research idea abstract generation

> Thought: The research idea focuses on developing a robust framework for detecting AI-generated content across multiple modalities by leveraging cognitive biases and advanced fusion techniques. The experimental steps outline a systematic approach to dataset collection, model development, fusion mechanism implementation, domain adaptation, evaluation, and deployment. The abstract should succinctly capture the research objectives, methods, expected results, and conclusions while maintaining clarity and precision.
> Abstract:
> {"Title": "Contrastive Meta-Style Adversarial Fusion Network for Robust Detection of AI-Generated Content",
> "Abstract": "This research proposes the Contrastive Meta-Style Adversarial Fusion Network (CMSAF-Net), a novel framework designed to detect AI-generated content across text, audio, and visual modalities by leveraging cognitive biases. The framework integrates modality-specific bias detection models, a contrastive hierarchical fusion mechanism, and a meta-style adversarial domain adaptation module. The methodology involves dataset collection, training of modality-specific models using advanced NLP, speech analysis, and computer vision techniques, and the development of a hierarchical fusion algorithm enhanced with contrastive learning. Additionally, meta-style adversarial training is employed to ensure generalization across diverse datasets. The framework is evaluated on a held-out dataset, with performance metrics including accuracy, precision, recall, and F1-score. Expected results indicate that CMSAF-Net will outperform existing unimodal and multimodal detection methods, providing a robust and generalizable solution for detecting synthetic media. The research concludes with the deployment of a user-friendly interface, aiming to mitigate the challenges posed by AI-generated content in misinformation campaigns."}

**Figure. A9.** An example for research idea abstract generation

```
Novelty: Whether the idea is creative and different from existing works on the topic, and brings fresh insights.
You are encouraged to search for related works online. You should consider all papers that appeared online
prior to 2024 as existing work when judging the novelty. A rating from 1 to 10. Here are the grading rules:
1. Not novel at all - there are many existing ideas that are the same
2.
3. Mostly not novel - you can find very similar ideas
4.
5. Somewhat novel - there are differences from existing ideas but not enough to turn into a new paper
6. Reasonably novel - there are some notable differences from existing ideas and probably enough to turn into
a new paper
7.
8. Clearly novel - major differences from all existing ideas
9.
10. Very novel - very different from all existing ideas in a very interesting and clever way
```
```
Feasible: How feasible it is to implement and execute this idea as a research project? Specifically, how feasible
the idea is for a typical CS PhD student to execute within 1-2 months of time. You can assume that we have
rich API resources, but only limited hardware resources. A rating from 1 to 10. Here are the grading rules:
1. Impossible: the idea doesn't make sense or the proposed experiments are flawed and cannot be implemented
2.
3. Very challenging: there are flaws in the proposed method or experiments, or the experiments require
compute/human resources beyond any academic lab
4.
5. Moderately feasible: It can probably be executed within the given time frame but would require careful
planning, efficient use of APIs or some advanced computational strategies to overcome the limited GPU
resources, and would require some modifications to the original proposal to make it work
6. Feasible: Can be executed within the given constraints with some reasonable planning
7.
8. Highly Feasible: Straightforward to implement the idea and run all the experiments
9.
10. Easy: The whole proposed project can be quickly executed within a few days without requiring advanced
technical skills
```
```
Excitement: How exciting and impactful this idea would be if executed as a full project. Would the idea change
the field and be very influential. A rating from 1 to 10. Here are the grading rules:
1. Poor: You cannot identify the contributions of this idea, or it's not interesting at all and you would fight to
have it rejected at any major AI conference
2.
3. Mediocre: this idea makes marginal contributions and is very incremental
4.
5. Leaning negative: it has interesting bits but overall not exciting enough
6. Learning positive: exciting enough to be accepted at a major AI conference, but still has some weaknesses
or somewhat incremental
7.
8. Exciting: would deepen the community's understanding or make major progress in this research direction
9.
10. Transformative: would change the research field profoundly and worth a best paper award at major AI
conferences
```

**Figure. A10.** Scoring rubric

**Note:** Some score values in the scoring rubric lack descriptions. This is because the granularity of the score levels is challenging to articulate in English. For specific details, please refer to the approach used in AI-Researcher[8].

---
[8] https://github.com/NoviScl/AI-Researcher

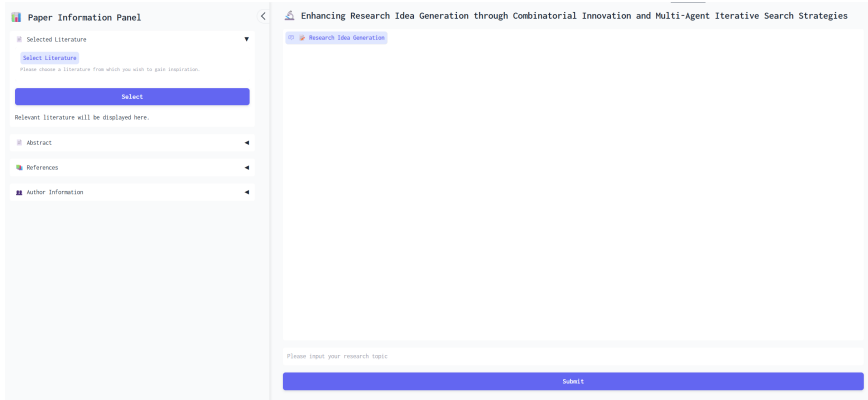

**Figure. A11.** Demonstration Interface

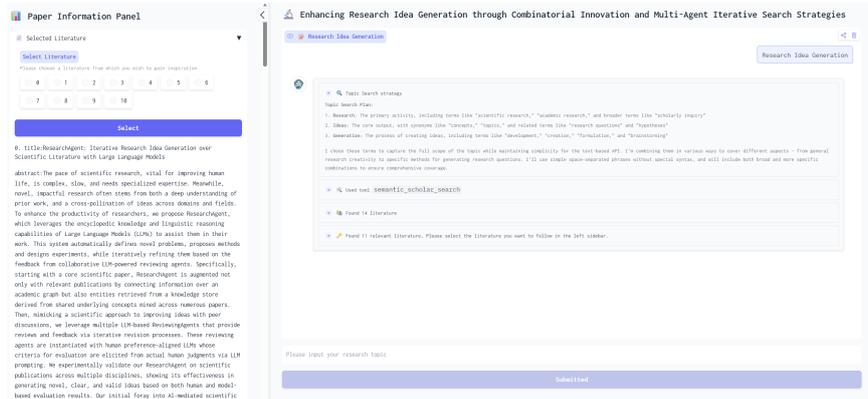

**Figure. A12.** Detailed Execution of the Literature Search Agent

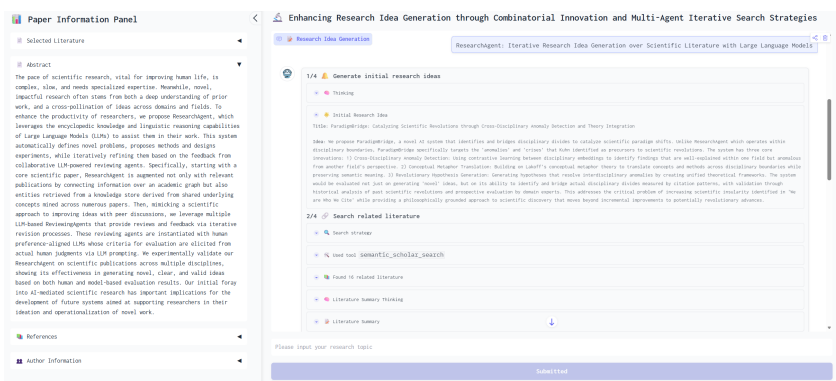

**Figure. A13.** Detailed Execution of Initial Research Idea Generation and Knowledge Search

**Figure. A14.** Virtual scientists refine research ideas

**Figure. A15.** Competition of Research Ideas

| Setting | Agents | Rounds | Paper Fetch (s) | Seed Idea (s) | Literature Search (s) | Agent Refinement (s) | Tournament (s) | Total (s) | Time per Idea (s) |
|---|---|---|---|---|---|---|---|---|---|
| Full pipeline, sequential | 3 | 2 | 12.58 | 48.34 | 94.17 | 106.57 | 0.00 | 261.67 | 87.22 |
| Full pipeline, parallel | 3 | 2 | 11.60 | 41.79 | 108.38 | 41.73 | 51.72 | 255.23 | 85.08 |
| Upstream only | 3 | 0 | 10.94 | 32.62 | 71.73 | 0.00 | 0.00 | 115.29 | — |

**Table. A16.** Stage-wise latency and Time-per-Idea under different execution settings.

> Title: CASPAC-SD: Dynamic Sparse Parameter Composition for Interference-Free Multi-Task Adaptation
>
> Idea: "Experiment": "Implementation Outline:\n1. **Base Model & Task Suite**: Select a pre-trained LLM (e.g., LLaMA-2-7B) as the frozen base model. Define a diverse set of 5-10 downstream tasks spanning different domains (e.g., code generation (HumanEval), medical QA (MedQA), mathematical reasoning (GSM8K), text summarization (CNN/DailyMail), and open-domain dialogue). Prepare datasets for all tasks.\n2. **Architecture Design**: \n - **Parameter Modulation Vectors (PMVs)**: Initialize a matrix of K PMVs (e.g., K=20, dim=256). Define a fixed, random projection basis (or a learned basis) to map each PMV to a sparse mask/scaling vector for each layer's weights.\n - **Context Router**: Implement a small transformer (e.g., 2 layers, 384 hidden dim) that takes the concatenation of the input's [CLS] token embedding (from the base model's encoder) and an optional task descriptor embedding. It outputs a K-dimensional softmax weight vector for PMV composition.\n - **Composition & Application**: The composed PMV = Σ (router_weight_i * PMV_i). This composed vector is projected to generate the final sparse modulation for the base model's weights.\n3. **Training Procedure**: \n - Use a mixed-batch training strategy: each mini-batch contains samples from multiple tasks (randomly sampled).\n - For each sample: (a) The context router produces weights, composes a PMV, and applies it to the base model for a forward pass; compute task loss L_task. (b) Perform a second forward pass using a fixed 'anchor' PMV (e.g., the mean of all PMVs); compute the self-consistency distillation loss L_scd as the KL divergence between the two output distributions, masked by the anchor model's confidence (entropy threshold).\n - Total Loss: L = L_task + β * L_scd + γ * (sparsity regularization on composed modulation).\n - Train only the PMV matrix and the router parameters.\n4. **Evaluation**:\n - **Primary Metrics**: Performance accuracy on each task's test set using the dynamically composed model. Compare to: (a) Individual LoRA fine-tuning per task (requiring K separate adapters), (b) Multi-task LoRA (single adapter trained on mixed data), (c) Vanilla multi-task fine-tuning (full fine-tuning), and (d) the base model zero-shot.\n - **Key Analysis**: \n - **Interference Test**: After training, evaluate on all tasks simultaneously using the dynamic router. Measure if performance on one task degrades when the model is frequently switched between tasks in a single session.\n - **Compositionality**: Test on 'compositional' tasks not seen during training (e.g., 'write a Python function that explains a medical concept') to see if the router can blend PMVs appropriately.\n - **Efficiency**: Measure inference latency and memory overhead compared to loading multiple LoRA adapters.\n - **Ablation Studies**: Ablate the self-consistency distillation loss and the sparsity penalty to show their role in stability and interference minimization.\nIdeal Outcomes: CASPAC-SD should achieve performance close to individually fine-tuned LoRAs on each task, while significantly outperforming multi-task LoRA and vanilla fine-tuning on the interference test (e.g., <2% performance drop when switching). The model should demonstrate emergent compositional ability on blended tasks. The parameter overhead (PMVs + router) should be far less than storing K separate LoRA adapters.

**Figure. A17.** Failure Case 1

> Title: ChronoAdapter: Continuous Temporal Adaptation of Language Models via Tensorized Hypernetworks
>
> Idea: "Experiment": "Implementation involves four key phases: 1) **Framework Construction**: Implement a base transformer (e.g., RoBERTa, LLaMA). Build the TCTH: a small MLP that outputs the core tensors for the TT-adapters. The TT-adapters will be inserted into the attention and FFN modules. The temporal encoder will use sinusoidal positional embeddings scaled for years/months. 2) **Dynamic Dataset Curation**: Construct a benchmark for evaluation, inspired by Papers 4, 8, and 9. We will create a temporal stream of tasks. For example, using Wikidata dumps (Paper 4), we create sequential question-answering tasks about entities, where facts are updated, added, or become obsolete over quarterly intervals from 2015-2023. We also incorporate a temporal reasoning dataset like TIME-News (Paper 9) for complex QA. 3) **Training Protocol**: We employ a continual learning setup. The model processes data in chronological order. For each mini-batch with timestamp *t* and task *k*, we: a) generate adapter parameters via the TCTH, b) perform a forward/backward pass on the backbone with these adapters active, c) update ONLY the TCTH parameters and the temporal encoder, using an optimizer that includes the conflict-alignment regularization from PS-LoRA. The backbone remains frozen. 4) **Evaluation & Analysis**: Ideal outcomes: (a) **Performance**: ChronoAdapter outperforms baselines like full fine-tuning per epoch, discrete LoRA per epoch, and a multi-task model with time-period heads on temporal QA accuracy, especially for recent time periods. (b) **Efficiency**: It uses <1% additional parameters per *temporal point* compared to the backbone, far less than storing separate adapters. (c) **Anti-Forgetting**: It maintains high accuracy on past temporal queries without rehearsal, demonstrated by evaluating on held-out test sets from all previous time periods. (d) **Extrapolation**: The model can make reasonable predictions for a future time period (e.g., 2024) by feeding future timestamps to the TCTH, showing it has learned a generalizable temporal dynamic. Ablation studies will confirm the necessity of the tensorized design, continuous time conditioning, and conflict regularization.",

**Figure. A18.** Failure Case 2

> TGA: Dynamic Graph-Structured Adapters for Lifelong Temporal Adaptation in Large Language Models
>
> Idea: "Experiment": "Implementation Outline:\n1. **Base Model & Adapter Initialization**: Select a base LLM (e.g., LLaMA-3-8B). Partition a historical diachronic corpus (e.g., 1M-News benchmark) into temporal chunks (e.g., yearly from 2010-2019). For each chunk, initialize a set of heterogeneous PEFT experts (e.g., 4 LoRA modules, 2 adapter layers per chunk) and pre-train them on their respective chunk's data for a target task (e.g., sentiment classification). This creates the initial pool of expert nodes.\n2. **Temporal-Semantic Relation Graph (TSRG) Construction**: Initialize a graph where nodes are experts. Node features combine the temporal centroid of their training chunk and a semantic signature (average task-specific embedding). Compute initial adjacency matrix A_ij using a composite kernel: A_ij = σ( β * exp(-|t_i - t_j|/τ) + (1-β) * cosine_sim(sem_i, sem_j) ), where σ is the sigmoid function. This graph is made learnable.\n3. **Dynamic Gating & Propagation Controller**: Design a lightweight controller network (e.g., a 2-layer MLP with a GAT layer). It takes the input text's [CLS] embedding (from the frozen base LLM) and a normalized timestamp embedding as input. This 'query' is used to compute initial node attentions. These attentions are then refined via one step of message passing on the TSRG to incorporate relational priors. The final node attention weights define a soft combination of expert outputs.\n4. **End-to-End Fine-Tuning**: On a validation period (e.g., 2020-2021 data), freeze the base LLM and jointly fine-tune the expert parameters, the TSRG adjacency weights, and the controller network. Use a standard task loss (e.g., cross-entropy) combined with a regularization loss on the TSRG to prevent over-smoothing.\n5. **Continual Learning Protocol**: Simulate a stream of new data (e.g., 2022-2023). For each new mini-batch:\n a. Use experience replay (Paper 10, 11) by storing a small core set of samples from past periods.\n b. Compute the Bayesian update (Paper 12) for the parameters of the top-k activated experts, treating their previous state as a prior.\n c. Update the TSRG edges connected to updated experts based on new similarity measures.\n6. **Evaluation**:\n - **Baselines**: Compare against (a) Static fine-tuned LLM, (b) TMSAF (seed idea ensemble), (c) Standard LoRA fine-tuning on all data, (d) BH-PEFT, (e) Experience Replay (ER) baseline.\n - **Metrics**: Primary: Task accuracy/F1 over sequential time periods (temporal robustness). Secondary: Parameter efficiency (% of trainable params vs. full fine-tuning), inference latency, forward transfer (on new tasks), and backward transfer (retention on old tasks).\n - **Ablation Studies**: Test importance of the TSRG structure vs. a flat mixture, the role of timestamp conditioning, and the contribution of Bayesian updates.\n - **Ideal Outcomes**: TGA matches or exceeds the accuracy of the ensemble-based TMSAF while using <5% of its storage footprint. It demonstrates superior resistance to catastrophic forgetting compared to standard PEFT methods and adapts to new trends faster than static models. The TSRG evolves to reflect learned temporal-semantic clusters of expertise.",

**Figure. A19.** Failure Case 3

> Phoenix-MoTE: Lifelong, Physics-Aware Adaptation of Foundation Models for Fleet-Wide Industrial Health Monitoring
>
> Idea: This framework enables a single pre-trained transformer model (initially trained on a diverse corpus of multi-sensor time-series data from bearings, gears, pumps, etc.) to adapt perpetually to new machines, new fault modes, and evolving degradation patterns within existing assets. The core innovation is a **Spatio-Temporal Operational Router (STOR)**, a lightweight network that takes two inputs: 1) a continuous temporal embedding derived from the asset's operational hours or timestamp, and 2) a spatial/operational context embedding derived from static metadata (e.g., machine type, load profile, RPM) and recent statistical features of the sensor stream (e.g., RMS, kurtosis). The STOR outputs sparse gating weights to select a small subset from a large, fixed pool of tensorized experts. Each expert is a **Physics-Informed TT-Adapter (PITA)**, inspired by LoTR (Paper 9) for parameter efficiency but augmented with inductive biases. Unlike purely data-driven adapters, each PITA incorporates a small, trainable component that models a specific physical prior (e.g., a differential equation kernel for vibration harmonics, a thermal dissipation model, a wear progression curve). These priors are encoded as structured, low-rank layers within the TT decomposition, allowing the expert to specialize in a particular fault physics (e.g., spalling, imbalance, lubrication failure) or operational regime (e.g., high-load, startup transient). The STOR learns to route sensor sequences from a specific machine at a specific point in its lifecycle to the most relevant physics-aware experts. For lifelong learning, we introduce **Operational Gradient Isolation with Memory Replay (OGI-MR)**. OGI penalizes interference between updates to experts specializing in different physical phenomena or asset types. Crucially, to combat catastrophic forgetting in the extremely low-data regime typical of PHM (where catastrophic faults are rare), OGI-MR integrates a minimal, dynamic memory buffer that stores \"prototypical\" sensor snippets for each learned fault/regime, using them for selective rehearsal. The significance is transformative for industrial AI: it provides a scalable, composable framework for building \"one model to monitor them all\"—a single foundation model that can be deployed across a heterogeneous fleet and continuously improve its diagnostic and prognostic accuracy over years, adapting to new equipment and newly observed failure modes without retraining from scratch. It moves PHM from static, model-per-asset approaches to a dynamic, lifelong learning paradigm. The potential impact spans manufacturing, energy, aerospace, and transportation, dramatically reducing unplanned downtime and maintenance costs."

Figure. A20. Failure Case 4